\documentclass[runningheads]{llncs}
\usepackage{graphicx}

\usepackage{tikz}
\usepackage{comment}

\usepackage{amsmath,amssymb} 
\usepackage{color}

\usepackage{blindtext}
\usepackage{float}
\usepackage{subcaption}
\DeclareMathOperator{\dis}{d}
\usepackage{mathtools}
\usepackage{capt-of}
\usepackage[pagebackref,breaklinks,colorlinks]{hyperref}
\newcommand{\etal}{\textit{et al.}}

\usepackage[accsupp]{axessibility}  

\begin{document}
\pagestyle{headings}
\mainmatter
\def\ECCVSubNumber{4910}  

\title{Implicit Field Supervision For Robust Non-Rigid Shape Matching} 

\titlerunning{Implicit Field Matching}
%
\author{Ramana Sundararaman \and
Gautam Pai \and
Maks Ovsjanikov}
\authorrunning{R.Sundararaman et al.}
%
\institute{LIX, École Polytechnique, IP Paris \\
\email{\{sundararaman, pai, maks\}@lix.polytechnique.fr}}

\maketitle

\begin{abstract}

Establishing a correspondence between two non-rigidly deforming shapes is one of the most fundamental problems in visual computing. Existing methods often show weak resilience when presented with challenges innate to real-world data such as noise, outliers, self-occlusion etc. On the other hand, auto-decoders have demonstrated strong expressive power in learning geometrically meaningful latent embeddings. However, their use in \emph{shape analysis} has been limited. In this paper, we introduce an approach based on an auto-decoder framework, that learns a continuous shape-wise deformation field over a fixed template. By supervising the deformation field for points on-surface and regularizing for points off-surface through a novel \emph{Signed Distance Regularization} (SDR), we learn an alignment between the template and shape \emph{volumes}. Trained on clean water-tight meshes, \emph{without} any data-augmentation, we demonstrate compelling performance on compromised data and real-world scans. \footnote{Our code is available at \url{https://github.com/Sentient07/IFMatch}}

\keywords{Non-rigid 3D Shape correspondence, Neural Fields}
\end{abstract}



\section{Introduction}
\label{sec:intro}
Understanding the relations between non-rigid 3D shapes through dense correspondences is a fundamental problem in computer vision and graphics. A common strategy is to leverage the underlying surfaces of shapes represented as triangle meshes. While recent advancements~\cite{sharp2021diffusionnet,eisenberger2021neuromorph} demonstrate near-perfect correspondence accuracies, they strongly rely on idealistic settings of clean input data, which unfortunately is far from typical 3D acquisition setups. The question of generalizability of non-rigid shape correspondence to artifacts such as noise, outliers, self-occlusions, clutters, partiality, etc. which are innate to general 3D scans, is largely unanswered.

On the other hand, 3D shape representations through neural fields~\cite{xie2021neuralfield} or learned implicit functions have been shown to achieve remarkable accuracy, flexibility and generative power for a wide range of shape and scene modeling tasks \cite{mescheder2019occupancy,chen2019learning,park2019deepsdf,sitzmann2019scene}. 
Unlike standard shape representations, learning implicit functions through a neural network allows one to capture continuous surfaces, while seamlessly adapting to changes in topology. Indeed, implicit surface representations not only allow to introduce an adaptive level of detail, but can also benefit from strong network regularization to control the desired resolution \cite{tancik2020fourfeat,sitzmann2020implicit}. As a result, although initial efforts have focused on using implicit representations primarily for generative modeling and shape recovery, several recent works have shown their utility in other tasks including differentiable rendering for image synthesis \cite{liu2020dist,takikawa2021neural},  part-level shape decomposition \cite{paschalidou2020learning}, modeling dynamic geometry \cite{niemeyer2019occupancy} and novel view synthesis~\cite{mildenhall2020nerf,Niemeyer2020GIRAFFE} among many others.

\begin{figure}[t]
\begin{center}
\includegraphics[width=1\linewidth]{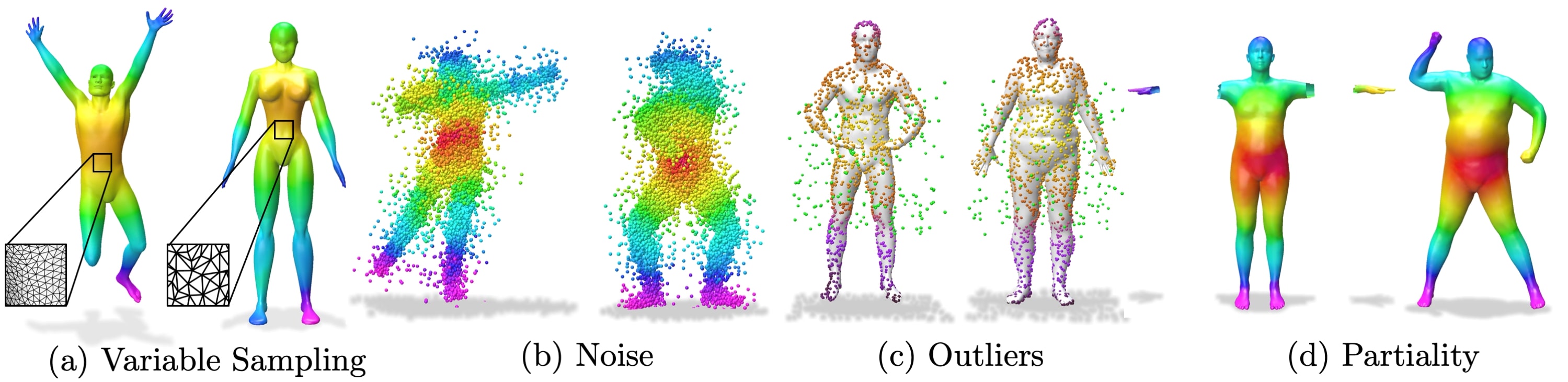}
\end{center}
\caption{Key advantages of our non-rigid shape correspondence pipeline: Our approach is extremely robust to common artifacts in 3D shapes like: (a) variations in sampling density, (b) significant noise, (c) cluttered outliers and (d) partiality.}
\label{fig:teaser}
\end{figure}
This flexibility of implicit surface representations, however, comes at a cost, especially in applications that involve multiple shapes, such as shape correspondence or comparison. Since the surface is defined as the zero-level set of a function, individual points are no longer easily identifiable. As a result, recent methods based on implicit surface representations that have aimed at shape alignment, try to model a warping field over an underlying template~\cite{deng2021deformed,zheng2021deep,Jiang2020ShapeFlowLD}, or between shape pairs~\cite{atzmon2021augmenting}. All of these works, however, primarily focus on deformations across nearby, sufficiently similar 3D shapes.

In this paper, we introduce an efficient method for establishing correspondences across \emph{arbitrary} non-rigid shapes, using neural field representations. To this end, we develop a new architecture based on the auto-decoder framework~\cite{park2019deepsdf}, that aims to recover a 3D deformation field between a fixed template and a target shape \emph{volume}. The key ingredient of our architecture is defining the shape-wise deformation field from the latent embedding, augmented with two effective regularizations. First, we regularize the deformation field for arbitrary points in space through a novel \emph{Signed Distance Regularization} (SDR). Second, we simultaneously condition the latent embedding to be compact and geometrically meaningful by learning a continuous Signed Distance Function (SDF) representation of the target shape. The resulting method is able to compute dense point-to-point correspondences between shapes while being extremely robust in the presence of varying sampling density, noise, cluttered outliers and missing parts as shown in Figure~\ref{fig:teaser}. To the best of our knowledge, ours is the first non-rigid correspondence method, based on neural field representation, that can be generalized to arbitrary shape categories such as articulated humans and animals.

Training on clean watertight meshes without any data-augmentation, we evaluate on a wide range of challenges across multiple benchmarks as well as real data captured by a 3D-scanner. Our approach shows compelling resilience to challenging artifacts and is more robust than existing point-based, mesh-based and spectral methods. In summary, our main contributions are: \textbf{(1)} We introduce an efficient approach based on the auto-decoder framework, capable of recovering a \emph{volumetric} deformation field to align a source and a target shape \emph{volumes}, even for significant non-rigid deformations. \textbf{(2)} We propose a novel way of regularizing the deformation of arbitrary points in space through the Signed Distance Regularization (SDR). \textbf{(3)} We perform rigorous evaluations by introducing challenges to existing benchmarks and on real-world data acquired by a 3D-scanner.

\section{Related Work}
\label{sec:relatedWork}

\subsection{Mesh-based Shape Correspondence}
There is a large body of literature on shape matching, for shapes represented as triangle meshes. We refer interested readers to recent surveys ~\cite{van2011survey,tam2012registration,biasotti2016recent,sahilliouglu2020recent} for a more comprehensive overview. Notable axiomatic approaches in this category are based on the functional maps paradigm  \cite{ovsjanikov2012functional,kovnatsky2013coupled,aflalo2013spectral,rodola2017partial,ezuz2017deblurring,burghard2017embedding}. Typically, these methods solve for near isometric shape correspondence by estimating linear transformations between spaces of real-valued functions, represented in a reduced functional basis. 
The conceptual framework of functional maps was further improved by learning-based formulations  \cite{litany2017deep,halimi2019unsupervised,roufosse2019unsupervised,Donati2020,eisenberger2020deep} that predict and penalize the map as a whole. Concurrently, recent advances in geometric deep learning have also tackled the correspondence problem by designing novel architectures for mesh and point cloud representation \cite{monti2017geometric,boscaini2016,masci2015geodesic,poulenard2018multi,wiersma2020cnns,lang2021dpc,zeng2020corrnet3d,eisenberger2021neuromorph}. Such methods typically treat the correspondence learning problem as vertex labelling, which is learned efficiently using the respective architectures. 

However, these methods that are predominantly based on mesh based representation of shapes are prone to sub-par performance when exposed to artifacts like sensitivity to variations in mesh discretization~\cite{sharp2021diffusionnet}, sampling, missing or occluded parts, noise and other challenges that are common in typical 3D acquisition setups. 

\subsection{Template Based Shape Correspondence}
Deforming a template to fit any given shape is a well-established technique in non-rigid shape registration~\cite{allen_articulated,allen_articulated_2}. The advent of learning-based skinning techniques~\cite{SMPL2015,ZuffiCVPR2017,ZuffiCVPR2018} enabled deformation of a fixed template to an arbitrary shape and pose by calibrating a fixed set of SMPL model parameters. The introduction of parametric models has opened the avenue for generating copious amounts of training data~\cite{varol17_surreal,groueix2018b,surreact} for data-driven methods. Such data-driven techniques have led to some seminal works in: 3D pose estimation~\cite{hmrKanazawa17,guler2018densepose,omran2018nbf}, digitizing humans~\cite{corona2021smplicit,Yu2019SimulCapS} and even model-based 3D shape registration for articulated humans~\cite{BogoCVPR2014,Pons-Moll-dyna,bogo_dfaust,Pons-Moll-Siggraph2017,NEURIPS2020_970af30e}. Most relevant to our work is LoopReg~\cite{NEURIPS2020_970af30e}, which proposes to diffuse SMPL parameters in space to learn correspondence. In contrast, our approach does not require any parametric models as priors and can be generalized across arbitrary categories.

On the other hand, there are techniques that learn a model-free deformation to align a fixed template to a target shape~\cite{groueix2018b,Deprelle2019LearningES,groueix19cycleconsistentdeformation,Wang_2019_CVPR}. Most notable among them is 3D-CODED~\cite{groueix2018b}, which learns to deform a fixed template mesh to a target shape. While this approach is succinct and well-founded, it requires significant amounts of training data to achieve optimal performance. Moreover, the deformation space is confined only to the surface of a mesh and can suffer from deformation artifacts. To ameliorate this, recent methods~\cite{deng2021deformed,zheng2021deep} have chosen to ``implicitly define the template''. However, their application in non-rigid shape matching is limited. 


\subsection{Neural Field Shape Representations}
Coordinate-based neural networks are emerging methods for efficient, differentiable and high-fidelity shape representations~\cite{park2019deepsdf,Atzmon_2020_CVPR,Chen2019LearningIF,sitzmann2020implicit,niemeyer2019occupancy,Genova2019LearningST,hao2020dualsdf,Wu_2020_CVPR,Zadeh2019VariationalA} whose fundamental objective is to represent zero level-sets using parameters of neural network. In its most general form~\cite{park2019deepsdf,sitzmann2020implicit,sitzmann2019scene}, these methods share two principal common goals - to perform differentiable surface reconstruction and to learn a latent shape embedding.  This has given rise to numerous applications especially in the field of generative 3D modelling~\cite{Tiwari2021NeuralGIFNG}, such as shape editing~\cite{hao2020dualsdf,takikawa2021nglod,Vasu2021HybridSDFCF}, shape optimization~\cite{Mezghanni_2021_CVPR,Gao_SDM} and novel view synthesis~\cite{mildenhall2020nerf,tancik2020fourfeat,Niemeyer2020GIRAFFE,Schwarz2020NEURIPS} to name a few. Most relevant to our work are DIF-Net~\cite{deng2021deformed} and SIREN~\cite{sitzmann2020implicit} which achieve shape-specific surface reconstruction through Hyper-Networks~\cite{ha2016hypernetworks}. However, leveraging the power of this representations in the domain of dense correspondence learning has so far been limited to nearly rigid objects~\cite{deng2021deformed,zheng2021deep,Genova2019LearningST,Jiang2020ShapeFlowLD}. 

\section{Method}
\textbf{Notation: } Throughout this manuscript, we use $\mathcal{S}$ to denote the target shape whose latent embedding is denoted by $\alpha_{\mathcal{S}} \in \mathbb{R}^{512}$ and $\mathcal{T}$ as the fixed template. $\mathcal{X}$ and $\mathcal{Y}$ denote an arbitrary pair of shapes between which we aim to find a correspondence. We let $\tilde{x} \in \partial \mathcal{S}$ be a point on the surface of the target shape $\mathcal{S}$, $x \in \mathbb{R}^{3}$ denotes a point in space and $\mathbf{\sigma}_{x}$ be its signed distance,  $\mathbf{\sigma}_{x} \coloneqq \dis(x, \partial \mathcal{S})$. We define $[\mathcal{S}] \coloneqq \{x \in \mathbb{R}^{3} | \sigma_{x} < \zeta \}$ to be the shape volume, which is the set of points sampled in space, in the vicinity of the shape surface $\partial \mathcal{S}$, with $\zeta$ being a constant. Analogously,  $[\mathcal{T}] \coloneqq \{t_{i} \in \mathbb{R}^{3} | \sigma_{t_{i}} < \zeta \}$ denotes the template volume. 

\begin{figure*}[t]
\includegraphics[width=1\linewidth]{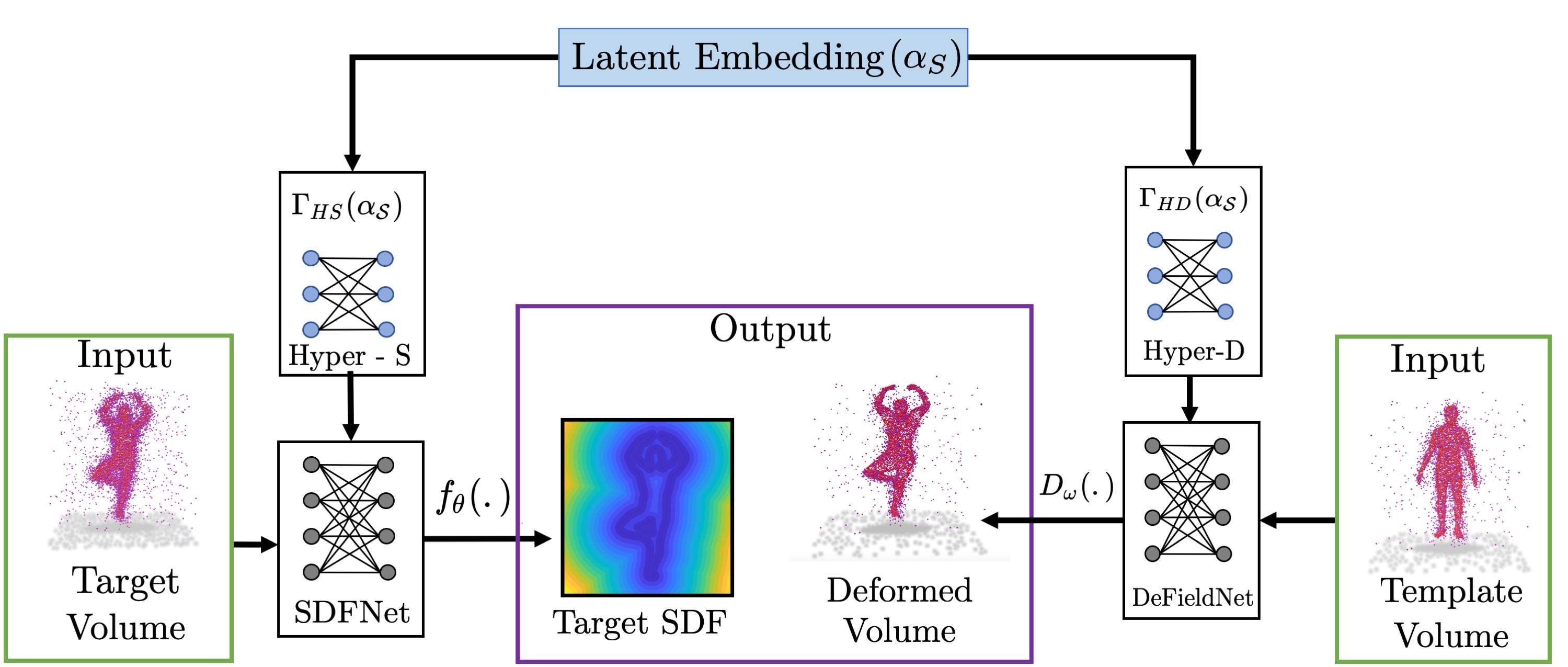}
\caption{Given a target shape volume (left) and a template volume (right) as input, DeFieldNet (Sec~\ref{sec:DeFieldNet}) aligns the template to target volume regularized by SDFNet (Sec~\ref{sec:SDFNet}). Shape-specific network weights are modeled by latent code (Sec~\ref{sec:overview}). Points sampled within volumes (input) are shown only for visualization purposes to emphasize that our network operates over the 3D domain.} 
\label{fig:overview}
\end{figure*}

\subsection{Overview}
\label{sec:overview}
Given a pair of shapes $\mathcal{X}$ and $\mathcal{Y}$, represented either as triangle meshes or point clouds, our goal is to estimate a point-wise map $ \Pi: \mathcal{X} \rightarrow \mathcal{Y}$. 
To this end, we learn a \emph{shape-specific} deformation field $D_{\omega} (.) : \mathbb{R}^3 \rightarrow \mathbb{R}^{3}$ which when applied to a fixed template volume $[\mathcal{T}]$, yields the target shape volume $[\mathcal{S}]$. Then, by using $D_{\omega} (.)$ to independently align $[\mathcal{T}]$ to $[\mathcal{X}]$ and $[\mathcal{Y}]$, we obtain correspondence between $\mathcal{X}$ and $\mathcal{Y}$ through nearest neighbor search. We stress that differently from previous data-driven works~\cite{groueix2018b,Deprelle2019LearningES} that align a template mesh to a target mesh, our approach aligns two \emph{volumes}. This is because, we observe that learning a volumetric alignment between arbitrary points in space naturally leads to a more robust map estimation as the deformation field is not constrained to an underlying surface defined by a mesh or a point cloud. While aligning on-surface points is straightforward in the supervised setting, aligning off-surface points is ill-posed. To this end, we propose a novel \emph{Signed Distance Regularization (SDR)} for constraining the change in the SDF brought about by the deformation field. Learning a continuous deformation field also allows us to impose useful smoothness and volume preservation constraints, for enhancing the regularity of the map. 

To make the deformation shape-specific, we learn a latent embedding $\alpha_{\mathcal{S}}$, which governs the parameters $\omega_\mathcal{S}$ of $D_{\omega_\mathcal{S}}(\cdot)$. We drop the subscript of $\omega$ for the sake of brevity. This latent embedding is learned following the auto-decoder framework~\cite{park2019deepsdf}. However, constructing an embedding based on the deformation field alone leads to topological inconsistencies as we discuss in the ablation studies (refer to Suppl). Therefore, we introduce a geometric prior to $\alpha_{\mathcal{S}}$ by learning a continuous Signed Distance Function (SDF) representation of the shape, resulting in two concurrent auto-decoder networks as shown in Figure~\ref{fig:overview}. On one side (left), we learn the continuous Signed Distance Function (SDF) of the target shape, which we refer to as \emph{SDFNet}. Simultaneously (right of Figure~\ref{fig:overview}), we learn a deformation field $D_{\omega}$ over $[\mathcal{T}]$ through \emph{DeFieldNet}. The parameters of our SDFNet $\theta \coloneqq \Gamma_{HS} (\alpha_{\mathcal{S}})$ and DeFieldNet $\omega \coloneqq \Gamma_{HD} (\alpha_{\mathcal{S}})$ are defined as two functions of the latent embedding, through Hyper-S and Hyper-D respectively. We perform an end-to-end training, to jointly learn the latent embedding $\alpha_{\mathcal{S}}$, through the gradients of SDFNet and DeFieldNet, similar to~\cite{ha2016hypernetworks,sitzmann2020implicit}. In summary, we learn a latent embedding by concurrently learning a deformation field over the template volume and the target shape's SDF. We stress that our main objective is to learn a plausible deformation field (via DeFieldNet) and the role of learning an implicit surface (via SDFNet) is to act as a geometric regularizer. 

\subsection{DeFieldNet}
\label{sec:DeFieldNet}
The main objective of DeFieldNet is to learn a smooth continuous shape-specific deformation field over the fixed template volume. We apply on surface supervision and off-surface regularization in order to deform the template volume $[\mathcal{T}]$ to the target shape volume $[\mathcal{S}]$.

\textbf{On Surface Supervision: }For two corresponding points $\tilde{x_{i}} \in \partial \mathcal{S}$ and $\tilde{t}_{i} \in \partial \mathcal{T}$, where $\Pi(\tilde{x_{i}}) = \tilde{t}_{i}$, our goal is to find a deformation $D_{\omega}: \tilde{t}_{i} \in \mathbb{R}^{3} \rightarrow \vec{v} \in \mathbb{R}^{3}, \hspace{2mm} s.t. \hspace{1mm} \tilde{t}_{i} + \vec{v} \approx \tilde{x_{i}}$. Thus, solving for the desirable deformation field amounts to optimising the following loss:
\begin{small}
\begin{equation}
\begin{aligned}
    \mathcal{L}_{\mathrm{surf}} &=\ \ \sum_{\tilde{x}_{i} \in \partial S}|| \tilde{x}_{i} - \hat{\tilde{x}}_{i}||_{2} \\
    \text{where}, \hat{\tilde{x}}_{i} &= D_{\omega}(\tilde{t}_{i}) + \tilde{t}_{i}
\end{aligned}
\label{eqn:ours_on_surface}
\end{equation}
\end{small}

\textbf{Signed Distance Regularization (SDR): } In addition to supervising the deformation of points on the surface, we also regularize the deformation field applied to arbitrary points in the template volume $t \in [\mathcal{T}] \in \mathbb{R}^{3}$. For this, we propose a \emph{Signed Distance Regularization} which \emph{preserves} the Signed Distance Function under deformation for points sampled close to the surface. More specifically, given signed distances: $\sigma_{t_{i}}, \sigma_{\hat{x}_{i}}$ of points $t_{i}$, $\hat{x}_{i}$ respectively where $\hat{x}_{i} = D_{\omega}(t_{i}) + t_{i}$, we require $\sigma_{t_{i}} \approx \sigma_{\hat{x}_{i}}$, for all points sampled closed to the surface. 

While $\sigma_{t_{i}}$ is available as a result of pre-processing, computing $\sigma_{\hat{x}_{i}}$ requires a continuous signed distance estimator as the SDF is measured w.r.t deformed shape. Therefore we perform \emph{discrete approximation} of the signed distance at any predicted point using Radial Basis Function (RBF) interpolation~\cite{Hardy1971}. For any $\hat{x}_{i} \in \mathbb{R}^{3}$, we first construct the RBF kernel matrix $\Phi$ as a function of its neighbors in the target shape volume $\mathcal{N}(\hat{x}_{i}) \in [\mathcal{S}]$.



\begin{equation}
\Phi_{\textit{ij}} \coloneqq \varphi(p_{i}, p_{j})=\sqrt{\varepsilon_{0} + ||p_{i} - p_{j}||^{2}}
\label{eqn:interpolant}
\end{equation}

Where, $\varphi$ is the radial basis function and $p_{i,j} \in \mathcal{N}(\hat{x}_{i})$. Assuming $\Delta = [ \sigma_1 \ldots \sigma_{K} ]^T$ to be the vectorized representation of the SDF values of neighbors, the estimated SDF $\hat{\sigma}_{\hat{x}_{i}}$ of $\hat{x}_{i}$ w.r.t deformed template $\mathcal{\tilde{T}}$ is given as:

\begin{equation}
\hat{\sigma}_{\hat{x}_{i}} = \varphi\left(\hat{x}_{i}\right) \Phi^{-1} \Delta 
\label{eqn:esitmatedSDF}
\end{equation}
We use shifted multiquadric functions as our RBF interpolant to avoid a singular interpolant matrix (refer to Suppl for more details). Therefore, our final SDF Regularization constraint can be written as:

\begin{equation}
\mathcal{L}_{SDR} = \sum_{t_{i} \in [\mathcal{T}]}||\operatorname{clamp}(\sigma_{t_{i}}, \eta) - \operatorname{clamp}(\hat{\sigma}_{\hat{x}_{i}}, \eta)||_{2}
\label{eqn:SDRConstraint}
\end{equation}

Where $\operatorname{clamp}(x, \eta):=\min (\eta, \max (-\eta, x))$ is applied to make sure that the penalty is enforced only to points close to the surface. We highlight that this clamping is necessary, since the change in SDF under a considerable non-rigid deformation may differ significantly for points far from the surface. 

\textbf{Smooth Deformation: }For the deformation field to be locally smooth, we ideally expect the flow vectors at neighboring points to be in ``agreement'' with each other. We enforce this constraint by encouraging the spatial derivatives to have minimal norm:

\begin{equation}
\mathcal{L}_{\mathrm{Smooth}}= \sum_{t_{i} \in [\mathcal{T}]}||\left.\nabla D_{\omega}\right(t_{i})\|_{2}
\label{eqn:our_smoothness}
\end{equation}

\textbf{Volume Preserving Flow: }Since a volume-preserving deformation field must be divergence-free, it must have a Jacobian with unit determinant~\cite{MaksSCA08}. 

\begin{equation}
\mathcal{L}_{\mathrm{vol}} =  \sum_{t_{i} \in [\mathcal{T}]} |\operatorname{det}(\nabla D_{\omega}(t_{i}))-1|
\label{eqn:our_volume}
\end{equation}

We use autograd to compute the Jacobian.
\subsection{SDFNet}
\label{sec:SDFNet}

Given a set of $N$ target shapes $\{\mathcal{S}_{0} \ldots \mathcal{S}_{N}\}$, our goal is to \emph{regularize} their latent embedding $\{\alpha_{\mathcal{S}_{0}} \ldots \alpha_{\mathcal{S}_{N}}\}$ through implicit surface reconstruction. We adopt the modified auto-decoder~\cite{sitzmann2020implicit} framework with sinusoidal $\mathcal{C}^{\infty}$ activation function as our SDFNet. Given $f_{\theta}(\cdot): x \in \mathbb{R}^{3} \rightarrow \sigma_{x} \in \mathbb{R}$ to be the function that predicts the Signed Distance for a point $x \in [\mathcal{S}]$, SDFNet's learning objective is given by,


\begin{small}
\begin{equation}
\begin{aligned}
\mathcal{L}_{\mathcal{S D} \mathcal{F}}=&\sum_{x \in [\mathcal{S}]} \left( |\left\|\nabla_{x} f_{\theta}(x)\right\|_{2} -1| + |f_{\theta}(x)-\sigma_{x}| \right) + \sum_{\tilde{x} \in \partial S} \left(1-\left\langle\nabla_{x} f_{\theta}(\tilde{x}), \hat{\mathbf{n}}(\tilde{x})\right\rangle\right)\\
&+\sum_{x \backslash \partial S} \psi(f(x))
\label{eqn:right_side}
\end{aligned}
\end{equation}
\end{small}

The first term penalizes the discrepancy in the predicted signed distance and enforces the Eikonal constraint for points in the shape volume. The second term encourages the gradient along the shape boundary to be oriented with surface normals. The last term applies an exponential penalty where $\psi \coloneqq \exp (- C \cdot|\sigma_{x}|), C \gg 0,$ for wrong prediction of $f_\theta(x) = 0$. 


\subsection{Training Objective: } In summary, the energy minimized at training time can be formulated as a combination of aforementioned individual constraints:

\begin{equation}
\mathcal{E}_{\mathrm{Train}} = \Lambda_{1} \mathcal{L_{SDF}} + \Lambda_{2} \mathcal{L}_{surf} +\Lambda_{3} \mathcal{L_{SDR}} + \Lambda_{4} \mathcal{L}_{Smooth} + \Lambda_{5} \mathcal{L}_{vol}
\label{eqn:our_training_energy}
\end{equation}

Here, $\Lambda_{i}$ are scalars provided in Sec~\ref{sec:implementation}. The first term helps to regularize the latent space, while the other terms encourage a plausible deformation field.

\subsection{Inference}
At inference time, given $\mathcal{X}, \mathcal{Y}$ to be a pair of unseen shapes, our approach is three-staged. First, we find the optimal deformation function $D_{\omega}$ associated with $\mathcal{X}, \mathcal{Y}$ to deform $[\mathcal{T}]$. We solve for optimal parameters for our deformation field $\omega$ through Maximum-a-Posterior (MAP) estimation as:

\begin{equation}
\begin{aligned}
\alpha_{i} &= \underset{\alpha_{i}}{\mathrm{argmin }} \hspace{2mm} \Lambda_{1}  \mathcal{L_{SDF}} + \Lambda_{3} \mathcal{L_{SDR}} \\
\omega &\coloneqq \Gamma_{HD}(\alpha_{i})
\end{aligned}
\label{eqn:our_test_energy_1}
\end{equation}

Second, similar to~\cite{groueix2018b} we enhance the deformation field applied by minimizing the bi-directional Chamfer's Distance

\begin{equation}
 \mathbf{\alpha_{opt}} = \underset{\alpha_{i}}{\mathrm{argmin}} \ \ \sum_{\mathbf{\tilde{s}} \in \partial \mathcal{S}} \min _{\mathbf{\tilde{t}}_{i} \in \partial \mathcal{T}}\left|D_{\omega} (\mathbf{\tilde{t}}_{i})-\mathbf{\tilde{s}}\right|^{2}+
 \sum_{\tilde{t}_{i} \in \partial \mathcal{T}} \min _{\mathbf{\tilde{s}} \in \partial \mathcal{S}}\left| D_{\omega} (\mathbf{\tilde{t}}_{i})-\mathbf{\tilde{s}}\right|^{2} 
\end{equation}

Finally, we establish the correspondence between $\mathcal{X}, \mathcal{Y}$ through their respective deformed templates using a nearest neighbor search.

\subsection{Implementation details}
\label{sec:implementation}
Our two Hyper-Networks, SDFNet and DeFieldNet all use 4-layered MLPs with 20\% dropout. SDFNet uses sinusoidal activation~\cite{sitzmann2020implicit} while DeFieldNet uses ReLU activation. We fix $\Lambda_1=1, \Lambda_2=500$, $\Lambda_3=50$, $\Lambda_4=5$, $\Lambda_5=20$, namely the coefficients in Equation~\ref{eqn:our_training_energy}.  For a shape in a batch, we use 4,000 points for on-surface supervision Equation~\ref{eqn:ours_on_surface}. We use 8,000 points for SDF regularization in Equation~\ref{eqn:SDRConstraint} and $\eta=0.1$ after fitting all shapes within a unit-sphere. We provide additional pre-processing details in the Suppl.  

\label{sec:approach}
\section{Experiments}
\textbf{Overview: }In this section we demonstrate the robustness of our method in computing correspondences under challenging scenarios through extensive benchmarking. We perform our experiments across 4 datasets namely, FAUST~\cite{JingBCICP}, SHREC'19~\cite{melzi2019shrec}, SMAL~\cite{ZuffiCVPR2017} and CMU-Panoptic dataset~\cite{Joo_2017_TPAMI}. The first three are mesh based benchmarks and are well-studied in non-rigid shape correspondence literature. In addition, we introduce challenging point cloud variants of these benchmarks which will be detailed below. CMU-Panoptic dataset~\cite{Joo_2017_TPAMI}, on the other hand, consists of raw point clouds acquired from a 3D scanner. 

For evaluation, we follow the Princeton benchmark protocol~\cite{Kim2011} to measure mean geodesic distortion of correspondence on meshes. We perform evaluation on our point cloud variants by composing the predicted map to the nearest vertex point and measure the mean geodesic distortion~\cite{Kim2011}. On the CMU-Panoptic dataset~\cite{Joo_2017_TPAMI}, we measure the error on established key-points. We stress that across all experiments, while the evaluations are performed under challenging scenarios, our model is trained on clean water-tight mesh \emph{without any data-augmentation}. Across all tables, ``$\text{*}$'' denotes a method that requires a mesh structure and cannot be evaluated on point clouds. ``$\text{**}$'' refers to computational in-feasibility in evaluating a baseline.

\textbf{Baselines: } We compare our method against several shape correspondence methods which can be broadly categorized into four main classes - axiomatic, spectral learning, template based and point cloud learning (PC Learning). We use ZoomOut (ZO)~\cite{MelziZoomout}, BCICP~\cite{JingBCICP} and Smooth Shells (S-Shells) ~\cite{Eisenberger2020SmoothSM} as our axiomatic baselines. For spectral basis learning baselines, we use Geometric Functional Maps (GeoFM)~\cite{Donati2020} with the recent more powerful DiffusionNet~\cite{sharp2021diffusionnet} feature extractor and DeepShells (D-Shells)~\cite{eisenberger2020deep}. We use 3D-CODED (3DC)~\cite{groueix2018b}, Deformed Implicit Fields (DIF-Net)~\cite{deng2021deformed} and Deep Implicit Templates (DIT-Net)~\cite{zheng2021deep} as template based baselines. We use Diff-FMaps (Dif-FM)~\cite{marin2020correspondence}, DPC~\cite{lang2021dpc} and Corrnet~\cite{zeng2020corrnet3d} as our point cloud learning baselines. For a fair evaluation, we identically pre-train them according to their category for different experimental settings as mentioned in the respective sections. We provide more details on the hyper-parameters used for baselines in the Supplementary. 

\subsection{FAUST}
\textbf{Dataset: }FAUST~\cite{BogoCVPR2014} dataset consists of 100 shapes where evaluation is performed on the last 20 shapes. Recently, Ren.~\etal~\cite{JingBCICP} introduced a re-meshed version of this dataset and Marin~\etal~\cite{marin2020correspondence} proposed a non-isometric, noisy point cloud version. For our robustness discussion, we introduce two additional challenges on top of the aforementioned variants. First, complimentary to~\cite{marin2020correspondence}, we introduce a dense point cloud variant consisting of 45,000 points perturbed with Gaussian noise. Second, we introduce 10\% clutter points by random sampling of points in space. In summary, we perform evaluation on (1) Re-meshed shapes~\cite{JingBCICP}, (2) Non-isometric noisy point cloud (NI-PC)~\cite{marin2020correspondence}, (3) Dense point clouds with noise (De-PC) and (4) Clutter.

\textbf{Baselines: } We train our model and all data-driven methods on the first 80 meshes of the FAUST dataset. All baseline methods are trained using the publicly available code, following the configuration stipulated by the respective authors.

\begin{figure}[H]
    \centering
    \includegraphics[width=0.8\linewidth]{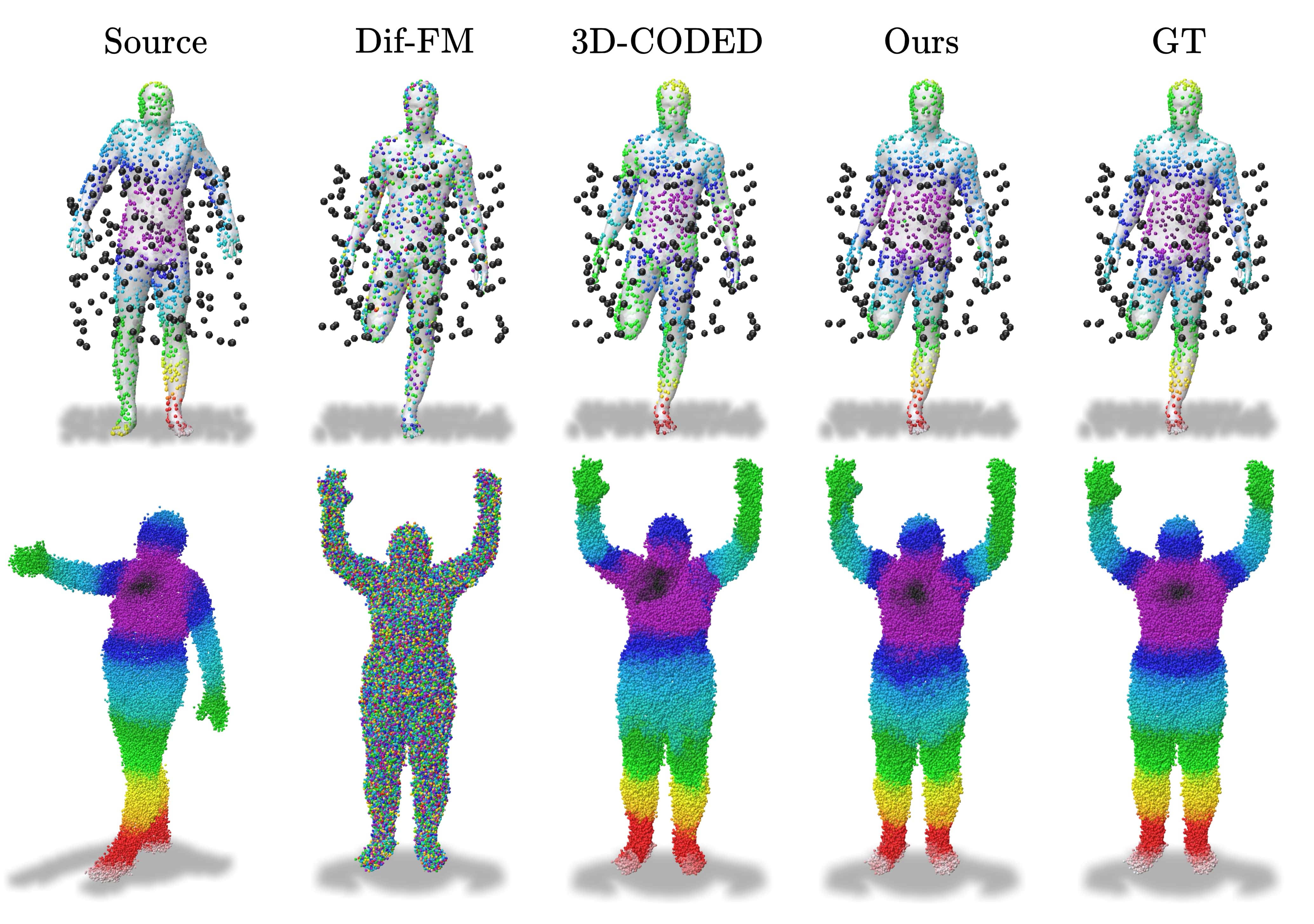}
    \caption{Correspondence quality through color transfer on challenges we introduced to FAUST~\cite{JingBCICP}. $1^{st}$ Row: Point clouds corrupted with 10\% clutter shown in black. In contrast to baselines, our method shows strong resilience in the presence of clutter. $2^{nd}$ Row: Point cloud with 45k points and noise. }
    
    \label{fig:faust_qualitative}
\end{figure}

\begin{table}[H]
\centering
\resizebox{\linewidth}{!}{%
\begin{tabular}{|c|ccc|ccc|cc|cccc|}
\hline
Category &
  \multicolumn{3}{c|}{Axiomatic} &
  \multicolumn{3}{c|}{PC Learning} &
  \multicolumn{2}{c|}{Spectral Learning} &
  \multicolumn{4}{c|}{Template Based} \\ \hline
Method &
  \multicolumn{1}{c|}{\begin{tabular}[c]{@{}c@{}} BCICP\\\cite{JingBCICP} \end{tabular}} &
  \multicolumn{1}{c|}{\begin{tabular}[c]{@{}c@{}}  ZO\\\cite{MelziZoomout} \end{tabular} } &
  \begin{tabular}[c]{@{}c@{}} S-Shells \\\cite{Eisenberger2020SmoothSM} \end{tabular}  &
  \multicolumn{1}{c|}{\begin{tabular}[c]{@{}c@{}} Dif-FM \\\cite{marin2020correspondence} \end{tabular}} &
  \multicolumn{1}{c|}{\begin{tabular}[c]{@{}c@{}} DPC \\\cite{lang2021dpc} \end{tabular}} &
  \begin{tabular}[c]{@{}c@{}} CorrNet \\\cite{zeng2020corrnet3d} \end{tabular} &
  \multicolumn{1}{c|}{\begin{tabular}[c]{@{}c@{}} D-Shells \\\cite{eisenberger2020deep} \end{tabular}} &
  \begin{tabular}[c]{@{}c@{}} GeoFM \\\cite{Donati2020} \end{tabular} &
  \multicolumn{1}{c|}{\begin{tabular}[c]{@{}c@{}} 3DC \\\cite{groueix2018b} \end{tabular}} &
  \multicolumn{1}{c|}{\begin{tabular}[c]{@{}c@{}} DIF-Net \\\cite{deng2021deformed} \end{tabular}} &
  \multicolumn{1}{c|}{\begin{tabular}[c]{@{}c@{}} DIT-Net \\\cite{zheng2021deep} \end{tabular}} &
  Ours \\ \hline
Remesh~\cite{JingBCICP} &
  \multicolumn{1}{c|}{10.5} &
  \multicolumn{1}{c|}{6.0} &
  2.5 &
  \multicolumn{1}{c|}{34.0} &
  \multicolumn{1}{c|}{27.1} &
  28.1 &
  \multicolumn{1}{c|}{\textbf{1.7}} &
  2.7 &
  \multicolumn{1}{c|}{2.5} &
  \multicolumn{1}{c|}{21.0} &
  \multicolumn{1}{c|}{20.1} &
  2.6 \\ \hline
\begin{tabular}[c]{@{}c@{}}NI-PC \\ + Noise \cite{marin2020correspondence} ~\end{tabular} &
  \multicolumn{1}{c|}{11.5} &
  \multicolumn{1}{c|}{8.7} &
  * &
  \multicolumn{1}{c|}{6.6} &
  \multicolumn{1}{c|}{8.4} &
  25.2 &
  \multicolumn{1}{c|}{*} &
  31.3 &
  \multicolumn{1}{c|}{7.3} &
  \multicolumn{1}{c|}{14.6} &
  \multicolumn{1}{c|}{13.6} &
  \textbf{3.1} \\ \hline
\begin{tabular}[c]{@{}c@{}}De-PC \\ + Noise  \end{tabular}  &
  \multicolumn{1}{c|}{*} &
  \multicolumn{1}{c|}{*} &
  * &
  \multicolumn{1}{c|}{31.8} &
  \multicolumn{1}{c|}{**} &
  27.9 &
  \multicolumn{1}{c|}{*} &
  53.7 &
  \multicolumn{1}{c|}{9.1} &
  \multicolumn{1}{c|}{18.1} &
  \multicolumn{1}{c|}{18.0} &
  \textbf{4.1} \\ \hline
Clutter &
  \multicolumn{1}{c|}{*} &
  \multicolumn{1}{c|}{*} &
  * &
  \multicolumn{1}{c|}{17.7} &
  \multicolumn{1}{c|}{50.0} &
  51.1 &
  \multicolumn{1}{c|}{*} &
  52.2 &
  \multicolumn{1}{c|}{22.1} &
  \multicolumn{1}{c|}{14.7} &
  \multicolumn{1}{c|}{14.3} &
  \textbf{8.1} \\ \hline
\end{tabular}}
\caption{Quantitative results on FAUST-Remesh dataset and its variants reported as mean geodesic error (in cm) scaled by shape diameter. }
\label{tab:faust_quantitative}

\end{table}

\textbf{Discussion: } Our main quantitative results are summarized in Table~\ref{tab:faust_quantitative}. On the re-meshed shapes~\cite{JingBCICP}, our method demonstrates comparable performance with existing state-of-the-art methods. However, as we decrease the perfection of data, our method shows compelling resilience towards artifacts and consistently outperforms all the other baselines by a noticeable margin. It is also worthy to remark that among all baselines that we compare with, our method is the only one that is capable of providing reasonable (less than 10cm) correspondence in the presence of clutter points. We also show two qualitative examples on our newly introduced variant in Figure~\ref{fig:faust_qualitative}.

\subsection{SHREC'19}
\label{sec:SHREC19}
 
\begin{figure}[t]
\centering
\includegraphics[width=0.8\linewidth]{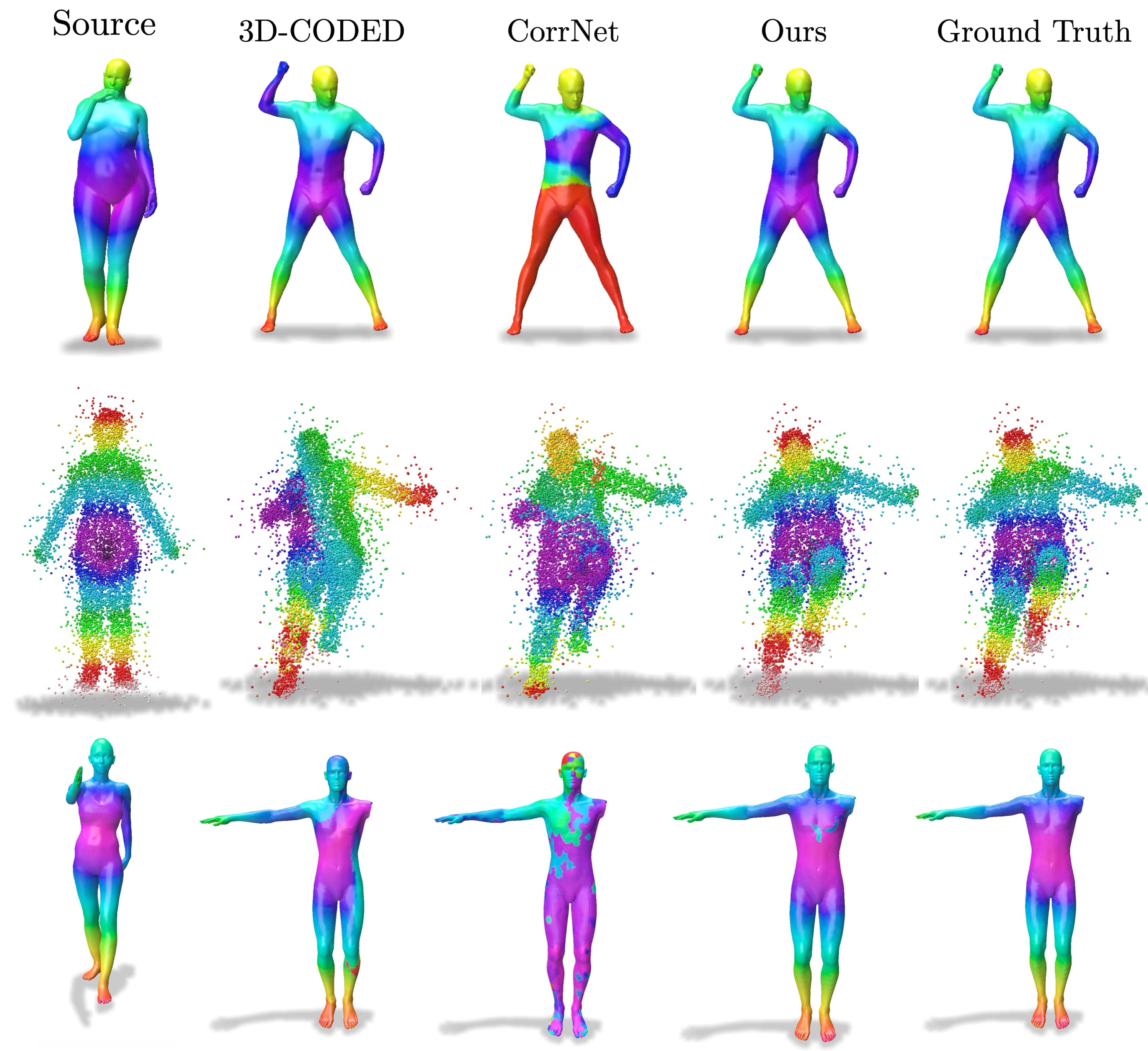}
\caption{Correspondence quality on SHREC'19~\cite{melzi2019shrec} and its variants. $1^{st}$ Row: Meshes. $2^{nd}$ Row: Point clouds with noise and outliers. $3^{rd}$ Row: Missing parts. Compared to baselines, our method exhibits strong resilience to artifacts.}
\label{fig:qual_shrec19}
\end{figure}

\textbf{Dataset: } SHREC'19~\cite{melzi2019shrec} is a challenging shape correspondence benchmark due to significant variations in mesh sampling, connectivity and presence of multiple connected components. It consists of 44 shapes and a total of 430 evaluation pairs. In addition, we introduce 3 challenging scenarios with different data imperfections. \textbf{Scenario 1: }We compare the meshes provided by Melzi~\etal~\cite{melzi2019shrec}. \textbf{Scenario 2:} We subsample the meshes to 10,000 points and introduce 20$\%$ outliers. \textbf{Scenario 3:} We further corrupt the surface information in Scenario 2 using Gaussian noise. \textbf{Scenario 4:} We introduce partiality in the form of missing parts, to a subset for a part-to-whole evaluation scheme~\cite{Rodol2016}.

\textbf{Baseline: }We pre-train all template based and point cloud learning baselines on 2,000 SURREAL shapes~\cite{varol17_surreal} including 10\% humans in bent poses~\cite{groueix2018b}. For our spectral basis learning baselines, we pre-train them on the training set of FAUST+SCAPE~\cite{SCAPE}, consisting of ${80 \choose 2} + {51 \choose 2}$ shape pairs, a setting which is demonstrated to be best suited for them~\cite{eisenberger2020deep,Donati2020}. We use Partial Functional Map (PFM)~\cite{Rodol2016} as an additional axiomatic baseline for Scenario 4.

\textbf{Discussion: } Quantitative results across 4 scenarios are summarized in Table~\ref{tab:shrec19-quantitative}. Our method demonstrates state-of-the-art performance across all variants of the SHREC'19 dataset and remains inert to imperfections in the data. While Smooth-Shells~\cite{Eisenberger2020SmoothSM}, is comparable to our approach in Scenario 1, it cannot be evaluated in other scenarios due to its strong dependence on spectral information. Moreover, even among template based methods, it is important to note that the supervised learning baseline 3D-CODED~\cite{groueix2018b} demonstrates significant decline in performance in the presence of outliers and noise. We posit that a well defined shape embedding, obtained by learning a volumetric mapping, plays a crucial role in our method's performance. Even among methods that construct a shape space through an auto-decoder framework, DIF-Net~\cite{deng2021deformed} and DIT~\cite{zheng2021deep}, are not reliable when presented with non-rigid shapes. Among point cloud learning methods, while DPC~\cite{lang2021dpc} shows comparable performance to our approach in Scenario 2, their performance declines in Scenario 3, when surface information is corrupted by noise. Furthermore, since DPC~\cite{lang2021dpc} depends on input point cloud resolution, it is infeasible to be evaluated in Scenarios 1 and 4. Finally, despite training on clean meshes with no missing components, the performance of our approach is unaffected by the partiality introduced in Scenario 4. We attribute our learning of \emph{volumetric alignment} coupled with off-surface regularization to be the reason behind robustness to missing components. We summarize this discussion by qualitatively depicting Scenarios 1, 3 and 4 in Figure~\ref{fig:qual_shrec19}, wherein, despite subsequently increasing artifacts, our method shows compelling resilience. Additional qualitative results in different poses are provided in the Supplementary.

\begin{table}[H]
\centering
\resizebox{\linewidth}{!}{%
\begin{tabular}{|c|c|c|c|c|c|c|c|c|c|c|c|}
\hline
\begin{tabular}[c]{@{}c@{}}Category\end{tabular} & \multicolumn{2}{c|}{Axiomatic} & \multicolumn{3}{c|}{PC Learning} & \multicolumn{2}{c|}{Spectral Learning} & \multicolumn{4}{c|}{Template Based} \\ 
\hline
\begin{tabular}[c]{@{}c@{}}\\Method\end{tabular} & \begin{tabular}[c]{@{}c@{}}S-Shells\\\cite{Eisenberger2020SmoothSM}\end{tabular} & \begin{tabular}[c]{@{}c@{}}PFM \\ \cite{Rodol2016} \end{tabular}& \begin{tabular}[c]{@{}c@{}}CorrNet\\\cite{zeng2020corrnet3d}\\\end{tabular} & \begin{tabular}[c]{@{}c@{}}DPC\\\cite{lang2021dpc}\\\end{tabular} & \begin{tabular}[c]{@{}c@{}}Diff-FM\\\cite{marin2020correspondence}\end{tabular} & \begin{tabular}[c]{@{}c@{}}GeoFM\\\cite{Donati2020}\end{tabular} &  \begin{tabular}[c]{@{}c@{}} D-Shells\\\cite{eisenberger2020deep}\end{tabular} & \begin{tabular}[c]{@{}c@{}}3DC\\\cite{groueix2018b}\\\end{tabular} & \begin{tabular}[c]{@{}c@{}}DIF-Net\\\cite{deng2021deformed}\end{tabular} & \begin{tabular}[c]{@{}c@{}}DIT-Net\\\cite{zheng2021deep}\end{tabular} & Ours \\ 
\hline
\begin{tabular}[c]{@{}c@{}}Scenario: 1\\ (Meshes)~\cite{melzi2019shrec}\end{tabular} & 7.6 &  N/A & 13.4 & ** & 29.6 & 11.7 & 15.2 & 9.2 & 14.9 & 41.4 & \textbf{6.5} \\ 
\hline
\begin{tabular}[c]{@{}c@{}}Scenario: 2\\ (Outliers)\end{tabular} & * &  N/A & 35.9 & 8.5 & 17.1 & 26.1 & * & 12.2 & 12.4 & 12.6 & \textbf{7.4} \\ 
\hline
\begin{tabular}[c]{@{}c@{}}Scenario: 3\\ (Outliers + Noise)\end{tabular} & * & N/A & 36.0 & 11.5 & 16.7 & 27.8 & * & 14.4 & 36.2 & 12.5 & \textbf{7.7} \\
\hline
\begin{tabular}[c]{@{}c@{}}Scenario: 4\\(Missing parts)\end{tabular} & * & 52.4 & 23.5 & ** & 26.3 & 48.6 & 23.8 & 6.0 & 11.9 & 41.1 & \textbf{4.3} \\
\hline
\end{tabular}}
\caption{Quantitative results on 430 test set pairs of SHREC'19 dataset reported as mean geodesic error (in cm), scaled by shape diameter.}
\label{tab:shrec19-quantitative}
\end{table}

\subsection{SMAL}
\textbf{Dataset: }In this section, we show the generalization ability to \emph{inter-class} non-rigid shape correspondence among to \emph{non-human} shapes. To this end, we use the SMAL dataset~\cite{ZuffiCVPR2017}, a parametric model that consists of 5 main categories of animals. We construct the training set by sampling 100 animals per each category. For correspondence evaluation, we generate 20 new shapes consisting of 4 animals per category, resulting in 180 \emph{inter-class} evaluation pairs. We relax the degrees of freedom for selected joints while generating the test-set to introduce new poses, unseen in the training set. In addition, we introduce partiality to this dataset in the form of multiple connected components. 

\textbf{Baseline Settings: }We train all template based methods, including ours, on the aforementioned 500 training shapes. For our method and 3D-Coded, which are supervised template based methods, we share the same animal template. Since spectral basis learning baselines learn correspondence pairwise, we train all data-driven spectral methods on ${100 \choose 2}$ shapes with 20 animals per-category.
\begin{figure}[H]

\includegraphics[width=1\linewidth]{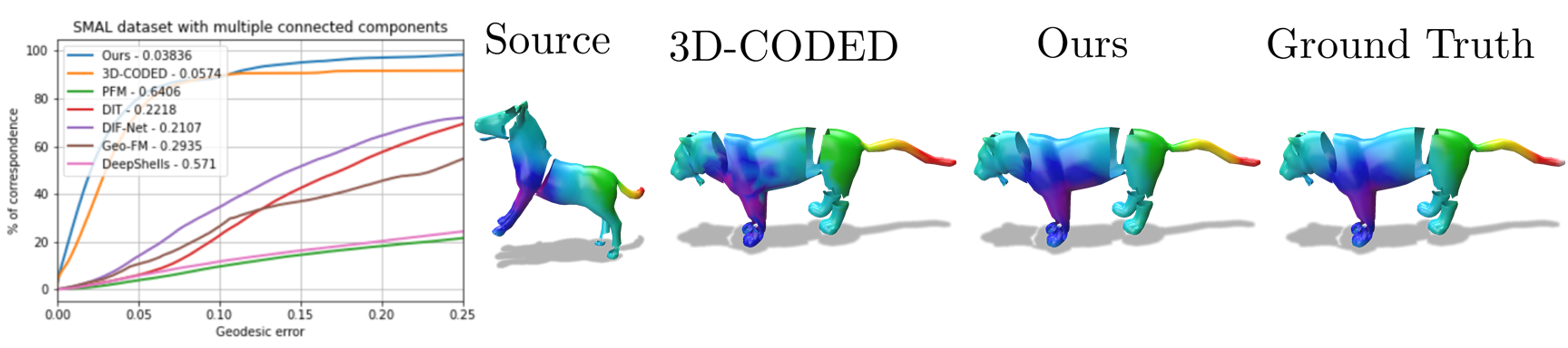}
\caption{Quantitative and qualitative inter-class correspondence on SMAL\cite{ZuffiCVPR2017} dataset. Our approach produces a smooth map, unaffected by partiality.}
\label{fig:qual_smal}
\end{figure}

\textbf{Discussion: }Our main quantitative and qualitative results are summarized in Figure~\ref{fig:qual_smal}. We observe that Geo-FM\cite{Donati2020,sharp2021diffusionnet} that is a representation agnostic method and Partial Functional Maps, an approach built to tackle partial non-rigid shape correspondence methods fail to establish reasonable correspondence. Our approach on the other hand, remains agnostic to shape connectivity arising from inter-class non-isometry and introduced partiality. Finally, our method surpasses the template-based baseline method, 3D-Coded by a considerable margin.

\subsection{CMU-Panoptic Dataset}
\textbf{Dataset: }In this section, we demonstrate the generalization ability of our approach to real-world sensor data. To that end, we use the CMU Panoptic~\cite{Joo_2017_TPAMI} dataset, which consists of 3+hrs footage of 8 subjects in frequently occurring social postures captured using the Kinect RGB+D sensor. This dataset consists of point clouds with noise, outliers, self-occlusions and clutter, allowing to evaluate correspondence methods on real-world data. We sample 200 shape pairs consisting of 3 distinct humans in 7 distinct poses. We measure non-rigid correspondence accuracy using the sparsely annotated anatomical landmark keypoints. More specifically, for each keypoint in the source, we consider 32 neighbors points and measure the disparity (as Euclidean distance, in cm) between their closest keypoint in the target and source.

\begin{figure}[ht]
\begin{center}
   \includegraphics[width=1\linewidth]{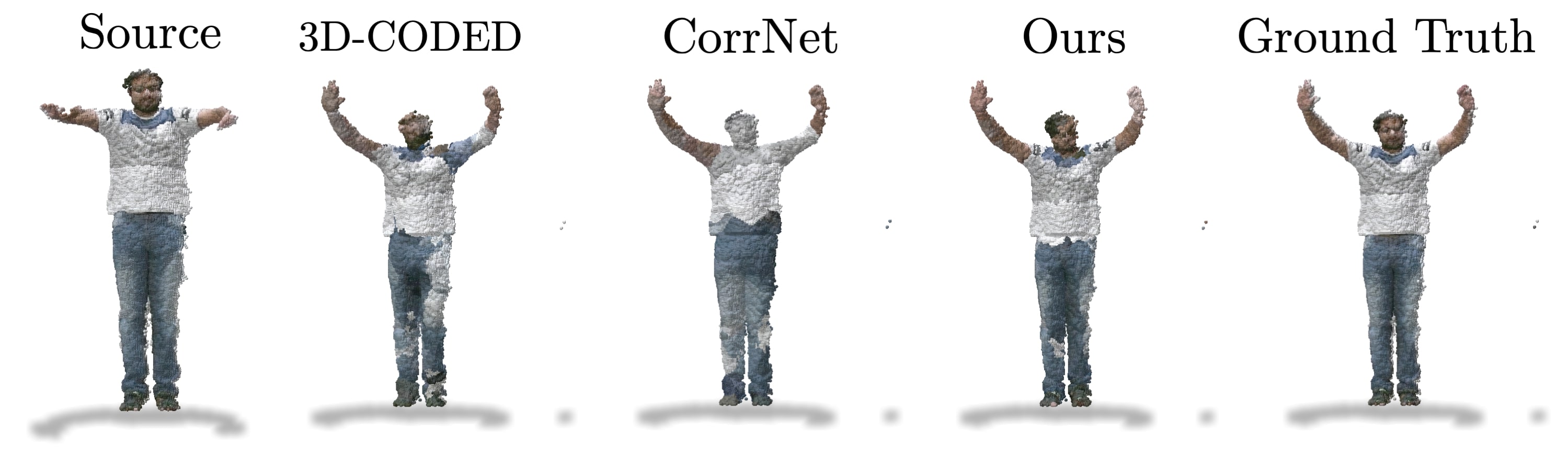}
\end{center}
\caption{Qualitative comparison using texture transfer on noisy point clouds from the CMU-Panoptic~\cite{Joo_2017_TPAMI} dataset.}
\label{fig:cmu-panoptic}

\end{figure}
\begin{table}[h]
\begin{center}
\resizebox{0.9\linewidth}{!}{%
\begin{tabular}{|l||l|l|l|l|l|l|l|}
\hline
Method & Dif-FM\cite{marin2020correspondence} & GeoFM\cite{Donati2020} & 3D-CODED\cite{groueix2018b} & DIF-Net\cite{deng2021deformed} & CorrNet\cite{zeng2020corrnet3d} & DPC~\cite{lang2021dpc} & Ours \\ \hline
Keypoint Error & 39.4 & 23.9 & 17.1 & 15.3 & 14.8 & ** & \textbf{8.5} \\ \hline
\end{tabular}}
\caption{Avg. Euclidean keypoint error (cm) for 200 test pairs, scaled by shape diameter. }
\label{tab:cmu-quantitative}
\end{center}
\end{table}

\textbf{Discussion: }In order for a fair evaluation of generalizability, we test all approaches using the trained model elaborated in Section~\ref{sec:SHREC19}. Quantitative results of keypoint errors are summarized in Table~\ref{tab:cmu-quantitative}. Our approach shows convincing performance in comparison to baselines, and more noticeably, it outperforms the conceptually closest supervised baseline, 3D-CODED~\cite{groueix2018b} by a twofold margin. We also show a qualitative example through texture transfer in Figure~\ref{fig:cmu-panoptic}, highlighting the efficacy of our approach in comparison to existing approaches on real-world data.

\label{sec:experiments_ramana}
\section{Conclusion, Limitations and Future work}
We presented a novel approach for robust non-rigid shape correspondence based on the auto-decoder framework. Leveraging its strong expressive power, we demonstrated the ability of our approach in exhibiting strong resilience to practical artifacts like noise, outliers, clutter, partiality and occlusion across multiple benchmarks. To the best of our knowledge, our approach is the first to successfully demonstrate the use of Neural Fields, which predominantly are used as generative models, to the field of non-rigid shape correspondence, generalizable to arbitrary shape categories.

Despite various merits, we see multiple avenues for improvement and possible future work. Firstly, our current framework of joint learning of latent spaces by continuous functions opens possibilities for local descriptor learning alongside purely extrinsic information. This can potentially lead to an unsupervised pipeline in contrast to our existing supervised method. Also, auto-decoder style learning approaches are not rotation invariant and conventional techniques like data-augmentation can prove costly in terms of training effort. Making Neural Fields rotational invariant is also an interesting future direction.
\label{sec:conclusion}

\textbf{Acknowledgements: } Parts of this work were supported by the ERC Starting Grants No. 758800 (EXPROTEA), the ANR AI Chair AIGRETTE. We thank Marie-Julie Rakotosaona and Nicolas Donati for their feedback in improving our manuscript, Riccardo Marin and Marvin Eisenberger for help with baselines and Qianli Ma for providing us CAPE Scans.

{\small
\bibliographystyle{splncs04}
\bibliography{egbib}
}

\end{document}


\pagestyle{headings}
\mainmatter
\def\ECCVSubNumber{4910}  

\title{Supplementary: Implicit field supervision for robust non-rigid shape matching} 

\titlerunning{Implicit Field Matching}
%
\author{Ramana Sundararaman \and
Gautam Pai \and
Maks Ovsjanikov}
\authorrunning{R.Sundararaman et al.}
%
\institute{LIX, Ecole Polytechnique, IP Paris \\
\email{\{sundararaman, pai, maks\}@lix.polytechnique.fr}}

\maketitle

In Section \ref{sec:sdf_regul}, we provide an elaborate illustration of the proposed Signed Distance Regularisation (SDR), followed by implementation details (Section~\ref{sec:additional_details}) and evaluation protocols (Section~\ref{sec:evaluation}). In Section~\ref{sec:ablation}, we perform an in-depth ablation study to quantitatively justify the efficacy of different components in our pipeline. In Section ~\ref{sec:more_exp}, we extend the robustness analysis by discussing the ability of our approach to endure varying noise levels and the impact of training data required to achieve optimal performance. Notably, we compare against methods trained with 100$\times$ more training shapes with data-augmentation and show that our approach, trained on a fraction of data is more robust. Finally, we conclude by discussing the known shortcomings of our method in Section~\ref{sec:limitations} and show more qualitative results over different challenging datasets in Section~\ref{sec:qual_results}. We emphasize that for this supplementary material, we \emph{do not} perform any additional parameter tuning or improve upon our reported results in the main submission.

\section{Signed Distance Regularization}
\label{sec:sdf_regul}

\begin{figure}[t]
\begin{center}
\includegraphics[width=\linewidth]{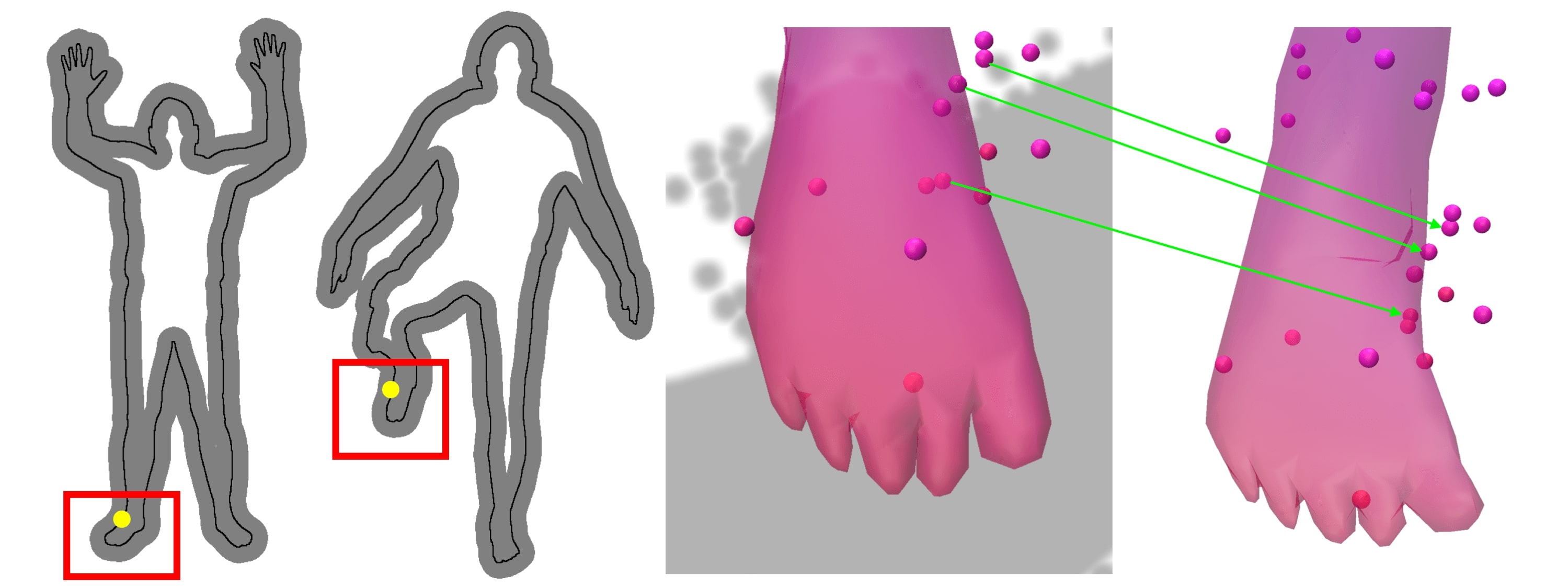}
\end{center}
\caption{Illustrating the key intuition behind Signed Distance Regularisation. (a) Given a point \emph{near} the surface in $[\mathcal{T}]$, (b) its corresponding point upon applying the deformation $D_\omega(\cdot)$ must \emph{approximately} have same the SDF. (c) and (d) : A particular case that shows SDF preserving the deformation field, owing to $\mathcal{L}_{\text{SDR}}$.}
\label{fig:sdf_preserve}
\end{figure}

To recall, we are given a template volume and target volume, denoted as $[\mathcal{T}]$, $[\mathcal{S}]$ respectively, which, we wish to align by learning a deformation field $D_\omega( \cdot )$. Let $t_{i} \in [\mathcal{T}]$ be a point sampled in the template volume and $x_{i} \in [\mathcal{S}]$ be a point in the shape volume. Let $\hat{x}_{i} \coloneqq t_{i} + D_{\omega}(\alpha_{i})$ be a point in space upon applying the deformation field $D_{\omega}(t_{i})$. We drop subscript $i$ for brevity. Let $\hat{\sigma}_{\hat{x}}$ be the signed distance of $\hat{x}$ in the shape volume $\hat{\sigma}_{\hat{x}} \coloneqq \dis (\hat{x}, \partial \mathcal{S})$ that we wish to estimate. Similarly, let $\sigma_{t}$, be the signed distance of $t$ in the template volume $\sigma_{t} \coloneqq \dis (t, \partial \mathcal{T})$. Then, our regularisation aims to preserve the SDF under the deformation $\sigma_{t} \approx \hat{\sigma}_{\hat{x}}$ as shown in Figure~\ref{fig:sdf_preserve}. \\

This regularisation is straightforward if $\hat{\sigma}_{\hat{x}}$ is known. However, in discrete settings, measuring $\hat{\sigma}_{\hat{x}}$ is not well-defined. To that end, we elaborate on the approximation technique using Radial Basis Function (RBF), introduced in the main paper. We begin by constructing the neighbourhood $\mathcal{N}\left(\hat{x}_{i}\right) = [x_{1} \ldots x_{K}]^{T}$ of $\hat{x}$ in the target shape volume $[\mathcal{S}]$. Please note that $\mathcal{N}(\hat{x}_{i})$ consists of points sampled in $[\mathcal{S}]$, whose SDF values are available as the result of pre-processing. Accordingly, let $\Delta = [\sigma_{1} \ldots \sigma_{K}]^{T}$ be the signed distance of $x_{j} \in \mathcal{N}(\hat{x}_{i}) \hspace{2mm} j \in [1, K]$. 

We used multiquadric kernel function as our interpolant, $\varphi(||p_{i}, p_{j}||) \coloneqq \sqrt{\varepsilon_{0} + ||p_{i} - p_{j}||^{2}}$ with $\Phi$ being the corresponding kernel matrix. Then, the \emph{interpolated} signed distance at the deformed point w.r.t target shape volume $[\mathcal{S}]$ is given as follows,

\begin{equation}
\hat{\sigma}_{\hat{x}}=\varphi(\hat{x}) \Phi^{-1} \Delta
\label{eqn:esitmatedSDF}
\end{equation}

The above equation has a solution \emph{iff} the kernel matrix $\Phi$ is invertible. For our choice of kernel function, it is easy to infer the following properties,

\begin{enumerate}
    \item $\varphi(||p_{i}, p_{j}||) \geq 0 \ \ \forall p_{i}, p_{j} \in \mathbb{R}^3$ 
    
    \item $\varphi(||p_{i}, p_{j}||) > 0 \ \ \forall p_{i}, p_{j} \in \mathbb{R}^{3}, \ \ s.t \ \ p_{i} \neq p_{j}$
\end{enumerate}

Therefore, $\varphi$ satisfies elementary properties of positive definiteness~\cite{Fleischmann1979} and hence our matrix $\Phi$ is always invertible. Furthermore, since we estimate $\hat{\sigma}_{\hat{x}}$ as a differentiable function of $\hat{x}$, the interpolation is differentiable w.r.t the input $t$ and can be used with auto-grad libraries. 

\section{Additional Implementation Details}
\label{sec:additional_details}
First, we provide additional details on pre-processing and training details concerning our method. Subsequently, we elaborate on the experimental setting of different baselines. 

\subsection{Pre-Processing}
\begin{figure}[t]
\begin{center}
\includegraphics[width=1\linewidth]{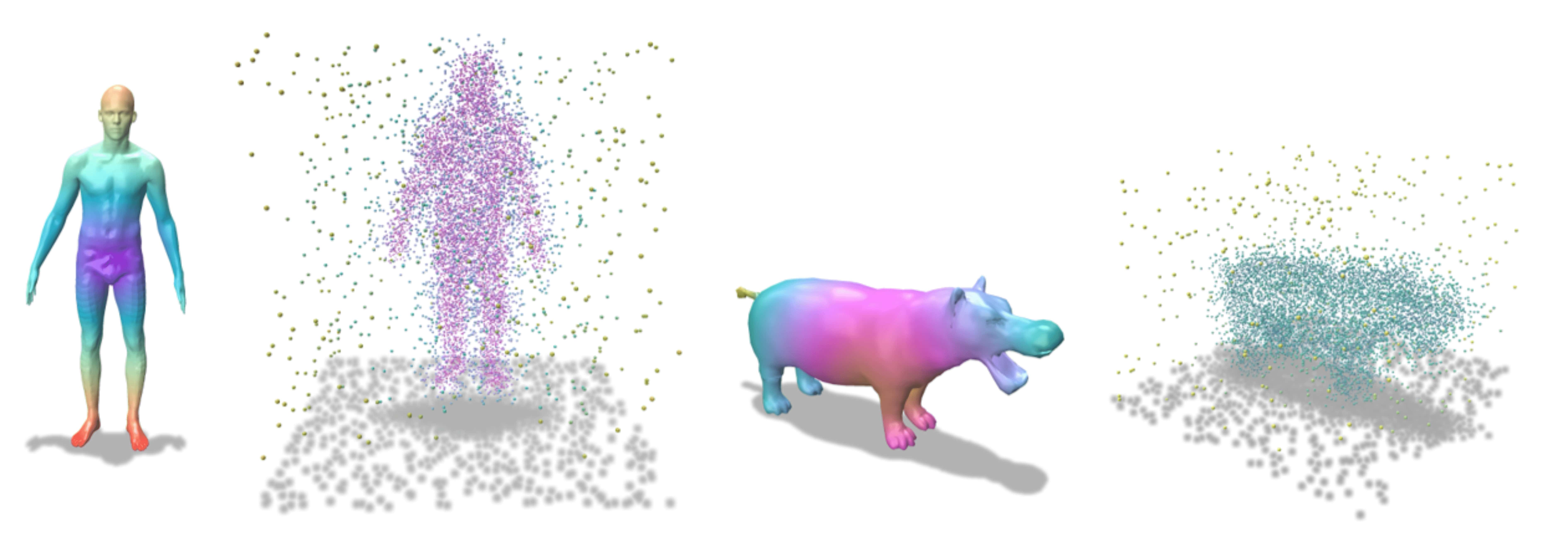}
\end{center}
\caption{Template meshes and respective \emph{volumes} for human and animal experiments. For visualization purposes, we depict 10,000 points sampled in the template volume.}
\label{fig:Templates}
\end{figure}

We start with a fixed template $\mathcal{T}$ and a set of shapes $\{\mathcal{S}_{0} \ldots \mathcal{S}_{N}\}$ with a known correspondence $\Pi$. We scale all shapes to fit within a unit sphere and align them along Y-axis, similar to previous works ~\cite{groueix2018b,zeng2020corrnet3d,lang2021dpc,SharmaO20}. This pre-processing step is performed for all baselines. We construct the shape volume $[\tilde{\mathcal{S}_{i}}]$ by sampling 400,000 points off-the surface of the shape. We perform this sampling aggressively close to the surface by displacing points sampled on the surface with a \emph{small} Gaussian noise. We estimate the signed distance of displaced points by placing 100 virtual laser scans of the shape from multiple angles, similar to ~\cite{park2019deepsdf}. This setup enables us to simultaneously compute surface normals for 20,000 points sampled on the surface of the shape. This pre-computed surface normals are used to enforce normal consistency prior in Equation 7 of the main paper. We perform this pre-processing independently and identically for template $\mathcal{T}$ to obtain $[\tilde{\mathcal{T}}]$ and $\sigma_{\mathcal{T}}$. As mentioned previously, we use two templates (analogously template volumes) across all experiments, namely, one human and one animal as depicted in Figure~\ref{fig:Templates}. 

\subsection{Training and Inference}
We train all our networks, namely Hyper-S, Hyper-D, SDFNet and DeFieldNet end-to-end and update the latent vector through back-propagation, a common practise in auto-decoder frameworks~\cite{park2019deepsdf,sitzmann2020implicit}. Although our two Hyper-Networks share the same input latent embedding, we \emph{stress} their weights are distinct and are initialized by the same latent vector. We use a learning rate of 1e-4 and train for 30 epochs with a batch size of 20. For experimental settings with no reliable information on ground truth SDF or normal information, we do not impose $\mathcal{L}_{SDR}$ and normal consistency terms of $\mathcal{L}_{SDF}$. In addition, at inference time, for point clouds, we consider SDF=0 for all points. We use the same coefficients as Sitzmann~\etal~\cite{sitzmann2020implicit} for our geometric regularization applied in Equation 7 of the main paper.  We train our network on an Nvidia A100 GPU for 12hrs requiring 2.3Gb of memory per-batch. We will release our code, pre-trained models and dataset variants introduced for full reproducibility.

\begin{figure}[t]
\begin{center}
\includegraphics[width=1\linewidth]{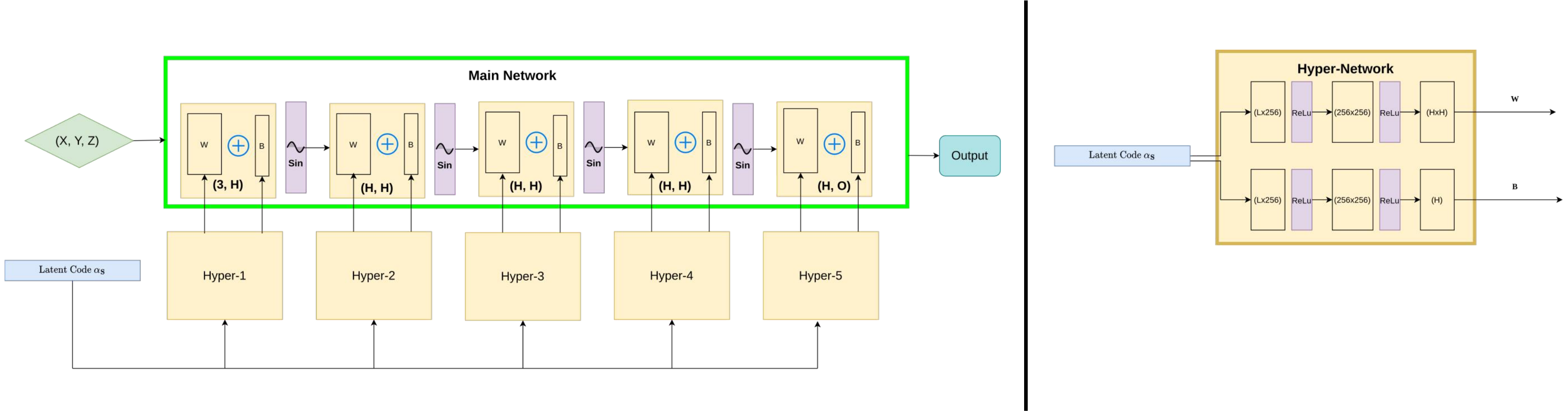}
\end{center}
\caption{Figure depicting the details of our SDFNet, DeFieldNet (left) and an individual Hyper-network block corresponding to Hyper-S and Hyper-D. Please refer to Section~\ref{subsec:arch} for more details.}
\label{fig:archDetails}
\end{figure}

\subsection{Network Architecture}
\label{subsec:arch}
A detailed depiction of our network's architecture is visualized in Figure~\ref{fig:archDetails}. The input coordinates (denoted as (X,Y,Z)) correspond to template volume for DeFieldNet and the target shape volume for SDFNet. ``H'' denotes the hidden dimension which we set to 256 for experiments with fewer than 1000 training shapes (c.f Section 4.1, 4.3 from the main paper) and 512 when using more than 2000 training shapes (c.f Section 4.2, 4.4 from the main paper). ``O'' denotes the output that lies in $\mathbb{R}^{3}$ for DeFieldNet and $\mathbb{R}^{}$ for SDFNet. Each Hyper-Net operates individually, predicting the weights and biases of corresponding layers of DeFieldNet and SDFNet respectively. An individual block of Hyper-Net is visualized in the right of Figure~\ref{fig:archDetails}, where each block denotes MLP followed by ReLU activation. 

\subsection{Run-time}
We report the run-time comparison between our approach and different baselines. For this, we consider one (top-performing, c.f. Table.2, main paper) baseline per category. Our observation is summarized in the Table~\ref{tab:runtime}. The run-time is measured per-pair, in seconds, averaged across 430 evaluation shapes of SHREC'19~\cite{melzi2019shrec}. While GeoFM outperforms the remaining approaches, this method is not built to handle point cloud inputs. On the other-hand, the second best performing axiomatic method S-Shells~\cite{Eisenberger2020SmoothSM} has a costly run-time.

\begin{table}[H]
\begin{center}
\resizebox{0.7\linewidth}{!}{%
\begin{tabular}{|l||l|l|l|l|l|}
\hline
Method & S-Shells~\cite{Eisenberger2020SmoothSM} & CorrNet~\cite{zeng2020corrnet3d} &  GeoFM~\cite{Donati2020} & 3DC~\cite{groueix2018b}   &   Ours \\ \hline
Run-time & 904.1  & 26.1      &   \textbf{4.2}         & 14.3      &    12.1 \\ \hline
\end{tabular}}
\end{center}
\caption{Comparison of inference run-time of different methods.}
\label{tab:runtime}
\end{table}

\subsection{Baselines}
We provide more details on various baselines used in our main paper. \textbf{Axiomatic: } First, we solve for a Functional Map~\cite{ovsjanikov2012functional} using 40 Eigenvalues on each shape with 20 Wave Kernel descriptors~\cite{aubryWKS} and refine the point-to-point map by spectral upsampling~\cite{MelziZoomout}, expanding the map size to 120x120. We refer to this as ZoomOut in our experiments. By introducing the orientation preservation operator, we optimize for the same map as before and refer to as BCICP~\cite{JingBCICP}. For Smooth Shells \cite{Eisenberger2020SmoothSM} and PFM \cite{rodola2017partial} we used the available code as is, using prescribed parameters in the respective papers. \textbf{Spectral Basis learning :} For GFM~\cite{Donati2020}, we used DiffusionNet~\cite{sharp2021diffusionnet} feature extractor consisting of 4 diffusion blocks with 128 dimensional layer-wise features. For all the point cloud based experiments, we computed the Point Cloud Laplacian~\cite{Sharp:2020:LNT} and used 33 Eigenvalues on each shape. For DeepShells~\cite{eisenberger2020deep}, we re-trained the author provided code without modifying the hyper-parameters. \textbf{Template learning :} We trained DIF-Net~\cite{deng2021deformed}, DIT~\cite{zheng2021deep} and 3D-CODED~\cite{groueix2018b} for 70 epochs, 2000 epochs and 100 epochs respectively. For 3D-CODED, we used the high-res template (230k vertices) and scaled the point cloud to match the spatial extents of template. \textbf{Point Cloud learning :} For DPC~\cite{lang2021dpc}, we used the author provided code and pre-trained model as their experimental setting are comparable to ours. For Corrnet3D~\cite{zeng2020corrnet3d} and Diff-FMap~\cite{marin2020correspondence}, we re-train on the same dataset as DPC using the author provided code for a fair evaluation. Additionally, Corrnet3D and DPC are trained only on 1024 input points. To scale the evaluation to arbitrary resolution, we follow the solution prescribed by the respective authors.

\section{Evaluation}
\label{sec:evaluation}

\subsection{Meshes}
We follow the Princeton benchmark protocol~\cite{Kim2011} for evaluating non-rigid shape matching accuracy for our mesh-based experiments. Given a predicted correspondence $\tilde{\Pi}$ and a ground truth correspondence $\Pi$ for shape $\mathcal{X}$, we measure the geodesic error as 
\begin{equation}
\varepsilon_{\mathcal{M}}(\tilde{\Pi}, \Pi) = \frac{\mathrm{d}_{\mathcal{G}}(\tilde{\Pi}, \Pi)}{\sqrt{\operatorname{area}(\mathcal{X})}}
\label{eqn:geod_error}
\end{equation}

In the partial setting, correspondence is evaluated only on the vertices that are present~\cite{Rodol2016}.

\subsection{Point Clouds}
Unlike for meshes, there is no universally accepted protocol for correspondence evaluation on point clouds. Hence, we created point cloud variants of SHREC’19~\cite{melzi2019shrec} and FAUST~\cite{JingBCICP} based on meshes from respective benchmarks for correspondence evaluation. We measure the correspondence error in two main steps. First, for each point in the source and target Point Clouds $x \in \ \mathcal{X},\  y \in \ \mathcal{Y}$, we construct \emph{Euclidean} maps $\mathcal{F}_{x}, \mathcal{F}_{y}$ that maps them to the nearest vertex in the underlying mesh. Given $\tilde{\Pi}$ and $\Pi$ to be predicted point-to-point map between point clouds and underlying mesh, we compose the two aforementioned maps to measure correspondence defined on mesh vertices as follows:

\begin{equation}
\varepsilon_{\mathcal{P}}(\tilde{\Pi}, \Pi) = \varepsilon_{\mathcal{M}}(\mathcal{F}_{\mathcal{Y}} \circ \tilde{\Pi} (\mathcal{X}), \Pi \circ \mathcal{F}_{\mathcal{X}}(\mathcal{Y}))
\label{eqn:PCEvaluation}
\end{equation}

Where $\varepsilon_{\mathcal{M}}$ is given in Equation~\ref{eqn:geod_error}.

\subsection{Key Point Evaluation}
We perform key point evaluation on the CMU-Panoptic dataset~\cite{Joo_2017_TPAMI}. This dataset consists of point clouds acquired from 3D-scans for which key-points annotations are available in the form 3D skeleton joints. There are in total 19 key-points following the Microsoft-COCO19 format~\cite{xiao2018simple}. For our evaluation, we consider these 19 key-points to be in correspondence, e.g. right-hip of two persons are in correspondence and measure the error in a \emph{small} key-point neighbourhood. More precisely, let $\kappa^{\mathcal{X}}_{i}$ and $\kappa^{\mathcal{Y}}_{j}$ be two key-points in correspondence, belonging to source $\mathcal{X}$ and target $\mathcal{Y}$ respectively. Let $\mathcal{N} : \kappa^{\mathcal{X}}_{i} \in \mathbb{R}^{3} \rightarrow X \in \mathbb{R}^{K \times 3}$ be a map that constructs a Euclidean neighbourhood around key-point $\kappa^{X}_{i}$ in the source such that $X \subset \mathcal{X}$. Here, $K$ denotes the size of neighbourhood and we set K=32 in our evaluation. Similarly, let $\mathcal{G} : Y \in \mathbb{R}^{K \times 3} \rightarrow \kappa^{\mathcal{Y}}_{j} \in \mathbb{R}^{3}$ be a map between points on target shape $Y \subset \mathcal{Y}$ to its nearest key-point. Considering $\Pi$ and $\tilde{\Pi}$ to be the ground truth map between key-points and predicted point-wise map respectively, the key-point error is measured as follows,

\begin{equation}
\varepsilon_{\mathcal{P}}(\tilde{\Pi}, \Pi) = \dis_{\mathcal{E}}(\mathcal{G}(\tilde{\Pi}(\mathcal{N}(\kappa_{i}^{\mathcal{X}}))), \Pi(\kappa_{j}^{\mathcal{Y}}))
\label{eqn:PCEvaluation}
\end{equation}

Where $\dis_{\mathcal{E}}$ is the Euclidean distance. 
\begin{figure*}[t]
\includegraphics[width=1\linewidth]{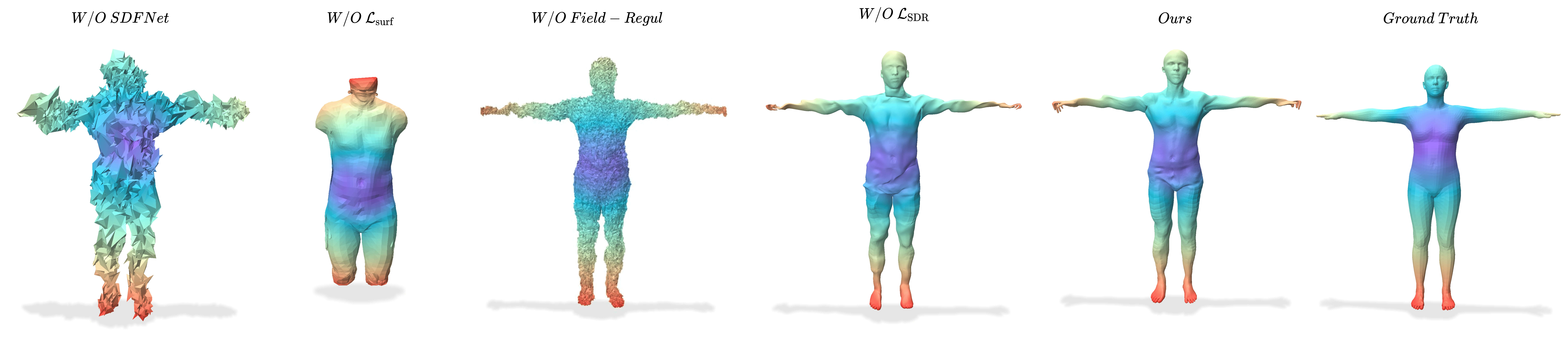}
\caption{Qualitative summary of our ablation study. Figures depict the reconstructed template mesh corresponding to different ablations. Inclusion of $\mathcal{L}_\mathcal{SDR}$ results in a smooth deformation field for points on and close to the surface.}
\label{fig:ablation}
\end{figure*}

\section{Ablation Studies}
\label{sec:ablation}
We justify the presence of each component in our network through an ablation study. We perform experiments on the FAUST-Remesh~\cite{JingBCICP} and SHREC'19~\cite{melzi2019shrec} datasets respectively. Training data and hyper-parameter details are in accordance with the main paper. We gauge the efficacy of each individual component by measuring correspondence accuracy of our network without different components listed herewith. 
\begin{enumerate}
    \item \textbf{w/o SDFNet: } The purpose of SDFNet is to regularize the latent embedding constructed from the deformation field through the gradients of DeFieldNet. We test the necessity of SDFNet by removing it. Our learning objective then becomes,
    $$
        \mathcal{L}_{\text{train}} = \mathcal{L}_\mathcal{SDR}+\mathcal{L}_{\text{surf }}+\mathcal{L}_{\text {vol     }}+\mathcal{L}_{\text {smooth }}
    $$
    Where we jointly optimize for shape latent space \emph{purely} based on deformation. Analogously, at test time we minimize the same objective without $\mathcal{L}_{\text {SDF }}$.

    \item \textbf{W/o $\mathcal{L}_{\text{surf}}$: } In the similar spirit of two conceptually similar prior works~\cite{zheng2021deep,deng2021deformed}, we try to reason for correspondence only through SDF representation. However, please note that different from the two aforementioned approaches, we use an explicitly defined template volume. Our new training objective is given by
    
    $$
        \mathcal{L}_{\text{train}} = \mathcal{L}_\mathcal{SDR}+\mathcal{L}_{\text {vol     }}+\mathcal{L}_{\text {smooth }}
    $$

    \item \textbf{Tr-Te W/o $\mathcal{L}_{\text{SDR}}$: }Our proposed SDR aims to regularize the deformation field by making \emph{preserve} signed distance under deformation. To understand its necessity, we remove $\mathcal{L}_{\text{SDR}}$ with the resulting loss that we minimize at training time,
    $$
        \mathcal{L}_{\text{train}} = \mathcal{L}_\mathcal{SDF}+\mathcal{L}_{\text {vol}}+\mathcal{L}_{\text {smooth}} +\mathcal{L}_{\text{surf }}
    $$
    
    Similarly, we remove $\mathcal{L}_\text{SDR}$ from the inference objective, corresponding to Equation. 9 in the main paper.

    \item \textbf{Te W/o $\mathcal{L}_{\text{SDR}}$: }While regularizing the deformation field at training time alone seem sufficient, it is also important to have a \emph{spatially consistent} deformation field at test time, i.e, the field must only map between level-sets. We hypothesize the highly non-convex nature of the optimisation to solve for a shape latent embedding to be a possible cause for this requirement. We empirically test this hypothesis by removing the $\mathcal{L}_\mathcal{SDR}$ term \emph{only} during inference. 

    \begin{equation*}
        \begin{aligned}
        \alpha_{i} &= \underset{\alpha_{i}}{\mathrm{argmin }} \hspace{2mm} \Lambda_{1}  \mathcal{L_{SDF}} \\
        \omega &\coloneqq \Gamma_{HD}(\alpha_{i})
        \end{aligned}
    \label{eqn:our_test_energy_woLSDR}
    \end{equation*}

Our training objective remains unchanged.  

    \item \textbf{W/O Field-Regul. :}Here, we try to understand different \emph{off-surface regularisations} applied to the deformation flow. such as $\mathcal{L}_{\text{smooth}},\  \mathcal{L}_{\text{SDR}},\  \mathcal{L}_{\text{vol}}$. Our training objective is therefore,
    
    $$
        \mathcal{L}_{\text{train}} = \mathcal{L}_\text{surf} + \mathcal{L}_\text{SDF}
    $$

Analogously, we remove the aforementioned terms at test-time.
    
     \item \textbf{W/O Opt: }Lastly, we remove Chamfer's Distance optimization (detailed in Equation 10 of the main paper) that is performed to enhance the deformation.  
\end{enumerate}

\textbf{Observation: }We summarize our quantitative results in Table~\ref{tab:ablation}. We make the following two main observations. First, while it might seem straightforward to learn a shape latent embedding only by supervising the deformation field, we observe a noticeable performance difference in correspondence accuracy across the two benchmarks SHREC'19~\cite{melzi2019shrec} and FAUST~\cite{JingBCICP} without our SDFNet. A possible explanation, coherent with our motivation, could be the efficacy of learning an implicit surface through the auto-decoder framework in providing \emph{geometrically meaningful} and compact latent embedding. Second, we also observe a discernible difference in performance with and without our proposed regularization, $\mathcal{L}_\mathcal{SDR}$. This observation is consistent with our hypothesis on the necessity to make the flow-field for points close to the surface spatially consistent. Moreover, making the deformation field preserve SDF also leads to a smoother reconstruction of template mesh as depicted in Figure~\ref{fig:ablation}. 

\begin{table}[b]
\centering
\resizebox{\textwidth}{!}{%
\begin{tabular}{|l|l|l|l|l|l|l|l|}
\hline
Experiment &
  W/O SDFNet &
  W/O Field-Regul &
  Tr-Te W/O $\mathcal{L}_{\text{SDR}}$ &
  Te W/O $\mathcal{L}_{\text{SDR}}$ &
  W/O$\mathcal{L}_{\text{surf}}$ &
  W/O Opt &
  Ours \\ \hline
\multirow{2}{*}{SHREC' 19} &
  \multirow{2}{*}{11.6} &
  \multirow{2}{*}{7.3} &
  \multirow{2}{*}{7.5} &
  \multirow{2}{*}{6.8} &
  \multirow{2}{*}{17.0} &
  \multirow{2}{*}{10.8} &
  \multirow{2}{*}{\textbf{6.5}} \\
      &      &     &     &     &      &     &     \\ \hline
FAUST & 14.8 & 4.9 & 3.7 & 3.6 & 26.8 & 5.0 & \textbf{2.6} \\ \hline
\end{tabular}%
}
\caption{Quantitative comparison of ablation study reported as mean geodesic error (in cm). Note that our model, using all components and losses leads to the lowest error}
\label{tab:ablation}
\end{table}

\section{Further robustness analysis}
We perform two additional experiments to consolidate our robustness discussion. First, we analyze the necessary training effort for our model to achieve optimal robustness in comparison to the closest supervised baseline, 3D-CODED~\cite{groueix2018b}. Second, we vary the levels of noise and clutter points for the experimental setting discussed before. Furthermore, in our second analysis, we compare our pre-trained model used in the main paper against baselines that were trained on \emph{$100 \times$ more training data}, i.e 230,000 shapes and with data-augmentation in the form of noise. We refer to such baselines as \emph{Oracle baselines} to the scope of this study. Subsequently, we demonstrate that our approach outperforms the baselines with a fraction of training data and without data-augmentation. 

\label{sec:more_exp}
\begin{figure}[t]
\includegraphics[width=1\linewidth]{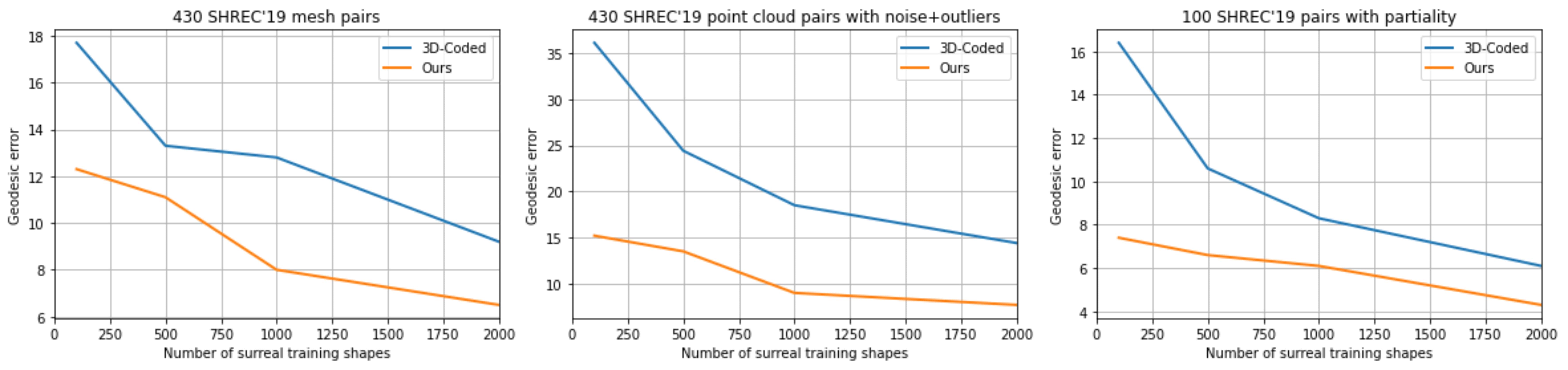}
\caption{Number of training shapes and corresponding geodesic error on SHREC19 and its variants. Since we perform partial (source) to full (target) shape matching, the evaluation in the last graph only consists of a subset.}
\label{fig:trainingdata_cmp}
\end{figure}
\subsection{Effect of training data}
We gradually increase the amount of training data and compare the correspondence accuracy across Scenario 1, Scenario 3 and Scenario 4 from Table. 2 in the main paper respectively. We construct four training sets consisting of 100, 500, 1000 and 2000 shapes from the SURREAL dataset~\cite{groueix2018b}. Our motivation behind this study is to demonstrate the efficacy of our approach in settings with paucity of training data. To this end, we compare with the closest supervised baseline, 3D-CODED~\cite{groueix2018b}, and show that in spite of being supervised, our approach needs significantly less training data, \emph{fractions} to be precise. Our approach and the baseline are trained with the same hyper-parameters as previously discussed. 

\textbf{Discussion: } Across three scenarios, we observe that our approach consistently outperforms the baseline irrespective of the number of samples in the training set as shown in Figure~\ref{fig:trainingdata_cmp}. Interestingly, in Scenario 3, where we introduce corruption to the data in the form of outliers, our approach achieves an error when trained on 100 shapes that is comparable to 3D-CODED trained on 2000 shapes. Finally, we observe over a two-fold improvement in performance in the partial setting with 100 training shapes. We posit that a \emph{stronger} conditioning of the latent embedding through SDF regularization and learning a \emph{volumetric map}, which is independent of the underlying geometry to be a possible reason behind this observation. 

\subsection{Comparison to Oracle baselines}

We compare our approach with three baselines, namely, 3D-CODED~\cite{groueix2018b}, Diff-FMaps~\cite{marin2020correspondence} and CorrNet3D~\cite{zeng2020corrnet3d}. The three aforementioned baselines are trained on 230k SURREAL shapes~\cite{groueix2018b} and thereby referred to as \emph{Oracle} baselines. However, we \emph{stress} again that we use our pre-trained network discussed in the main paper, trained on 2k SURREAL shapes. 

We compare our method to the baselines by varying the level of corruption to data, across experimental settings studied in the main paper. To that end, we further subdivide this study in two experimental settings. First, we consider the variant of FAUST consisting of point clouds with clutter points. Second, we evaluate on the variant of SHREC'19 involving point clouds with noise and outliers respectively. For the first case, we use 15\%, 25\% and 30\% clutter points in contrast to 20\% of the total points discussed in the main paper. Similarly, for the second case, we vary the standard deviation of the Gaussian noise added to the surface between 0.5\% and 25\%, in contrast to 10\% discussed in the main paper. Furthermore, for the second case, we add a \emph{stronger} Gaussian Noise with standard deviation $\sigma=0.1$ to 20\% of the points in the point cloud.

\begin{figure*}
\includegraphics[width=1\linewidth]{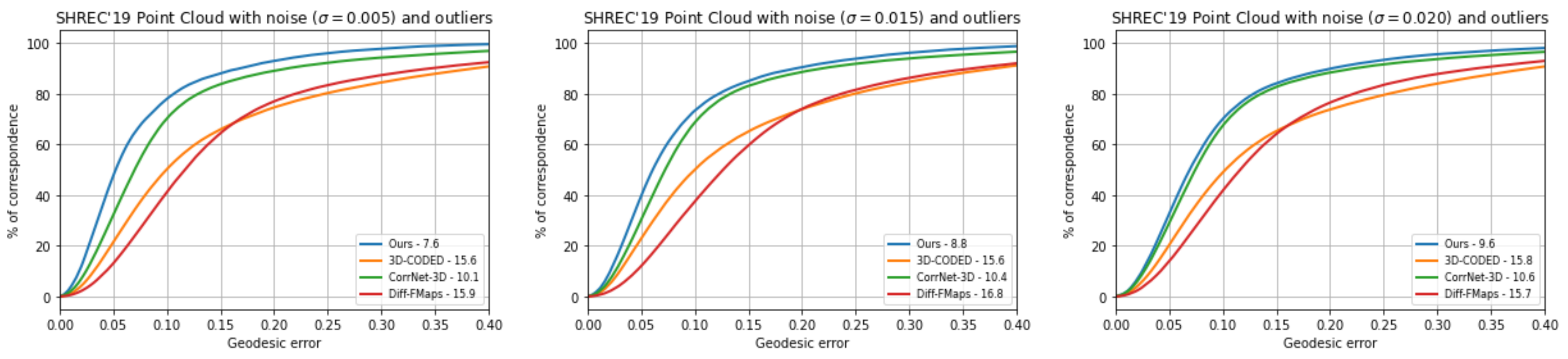}
\caption{Quantitative comparison for matching point clouds with varying levels of noise. Our method is trained on 2000 training shapes while all the \emph{Oracle} baselines are trained on 230k shapes. }
\label{fig:noise_shrec}
\end{figure*}

\begin{figure*}
\includegraphics[width=1\linewidth]{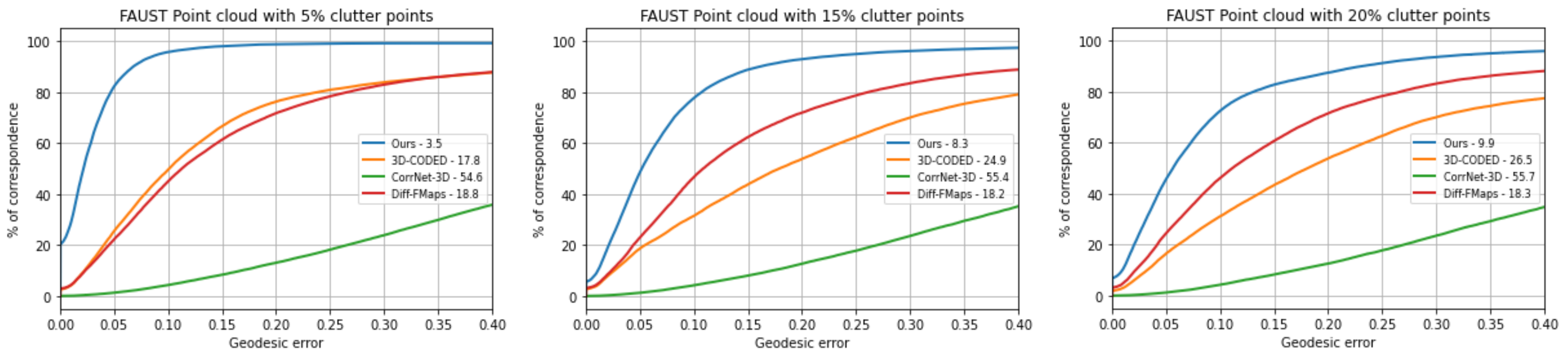}
\caption{Quantitative comparison between our method and different baselines for matching point clouds in the presence of varying levels of clutter points. Our method is trained on 2000 training shapes while all the baselines are trained on 230k shapes. }
\label{fig:clutter_faust}
\end{figure*}

\textbf{Discussions: }Our results are summarized quantitatively through geodesic accuracy graphs~\cite{Kim2011} in Figure~\ref{fig:noise_shrec} and Figure~\ref{fig:clutter_faust} respectively. Consistent with our observation in the main paper, our method shows high resilience towards noise and imperfection in data. Our aim of reducing the amount of noise is to show that performance of existing state-of-the-art methods rapidly degrades even in the presence of \emph{negligible} imperfection in the data.

\section{Qualitative results}
\label{sec:qual_results}
Finally, we show qualitative results across different benchmarks, namely the noisy point cloud variant of SHREC'19 mentioned in our main paper, additional qualitative examples of scanned point clouds from CMU Panoptic dataset~\cite{Joo_2017_TPAMI}, animal shapes from Deforming Things 4D~\cite{li20214dcomplete} and real-world scans of humans in clothing with registration artifacts from CAPE Scans dataset~\cite{CAPE:CVPR:20}.

\subsection{SHREC'19 Point clouds with outliers}
\begin{figure}[H]
\includegraphics[width=1\linewidth]{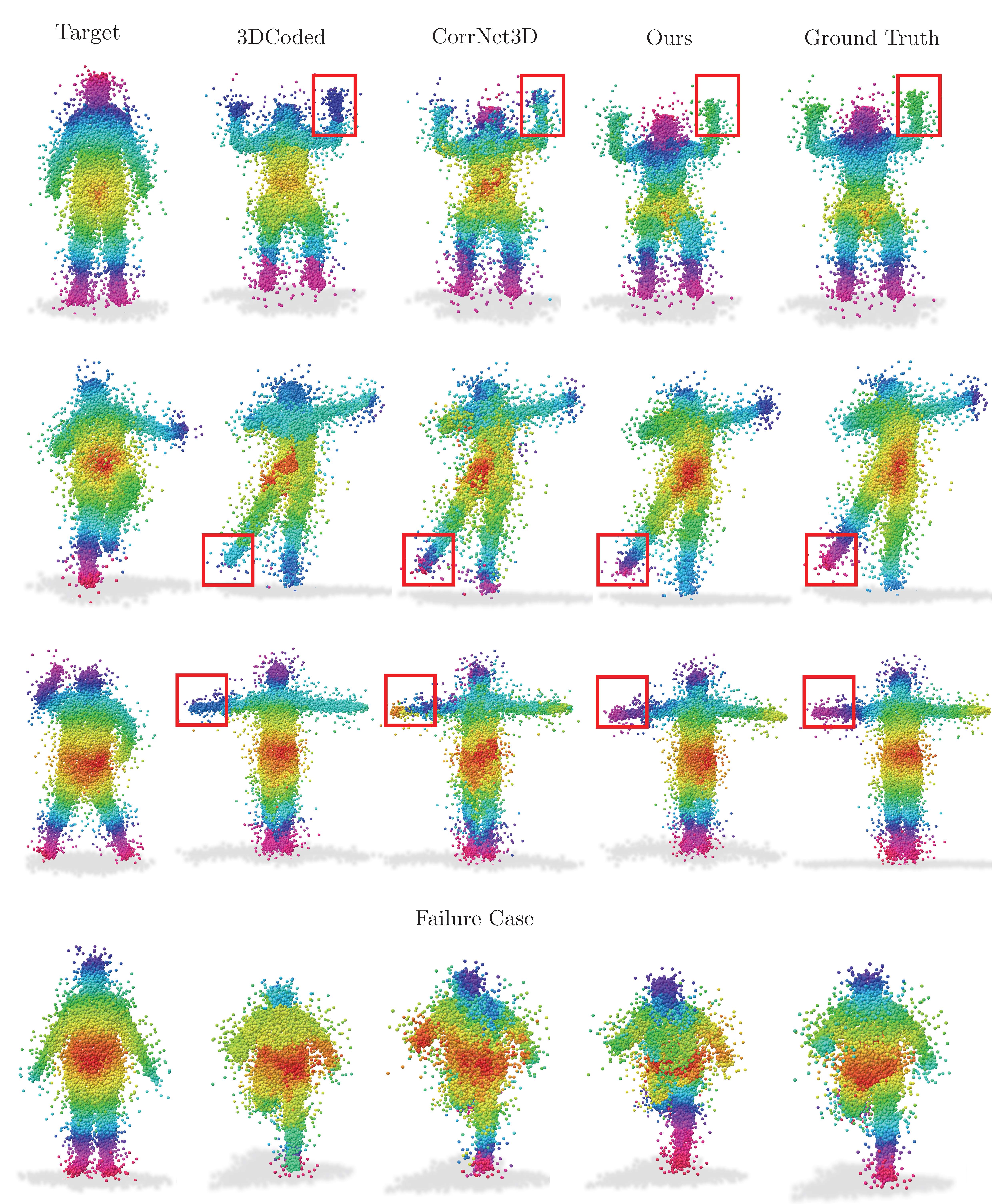}
\caption{Additional qualitative results of our approach and the baselines 3DCoded \cite{groueix2018b}, CorrNet3D \cite{zeng2020corrnet3d} on SHREC'19 point clouds with outlier introduced in the main paper. For ease of observation, we highlight stark differences in map quality in red. In the final row, we also report a failure case of our approach.}  
\label{fig:qual_additional_shrec19}
\end{figure}

\subsection{Point Clouds from CMU Panoptic dataset}
\begin{figure}[H]
\includegraphics[width=0.9\linewidth]{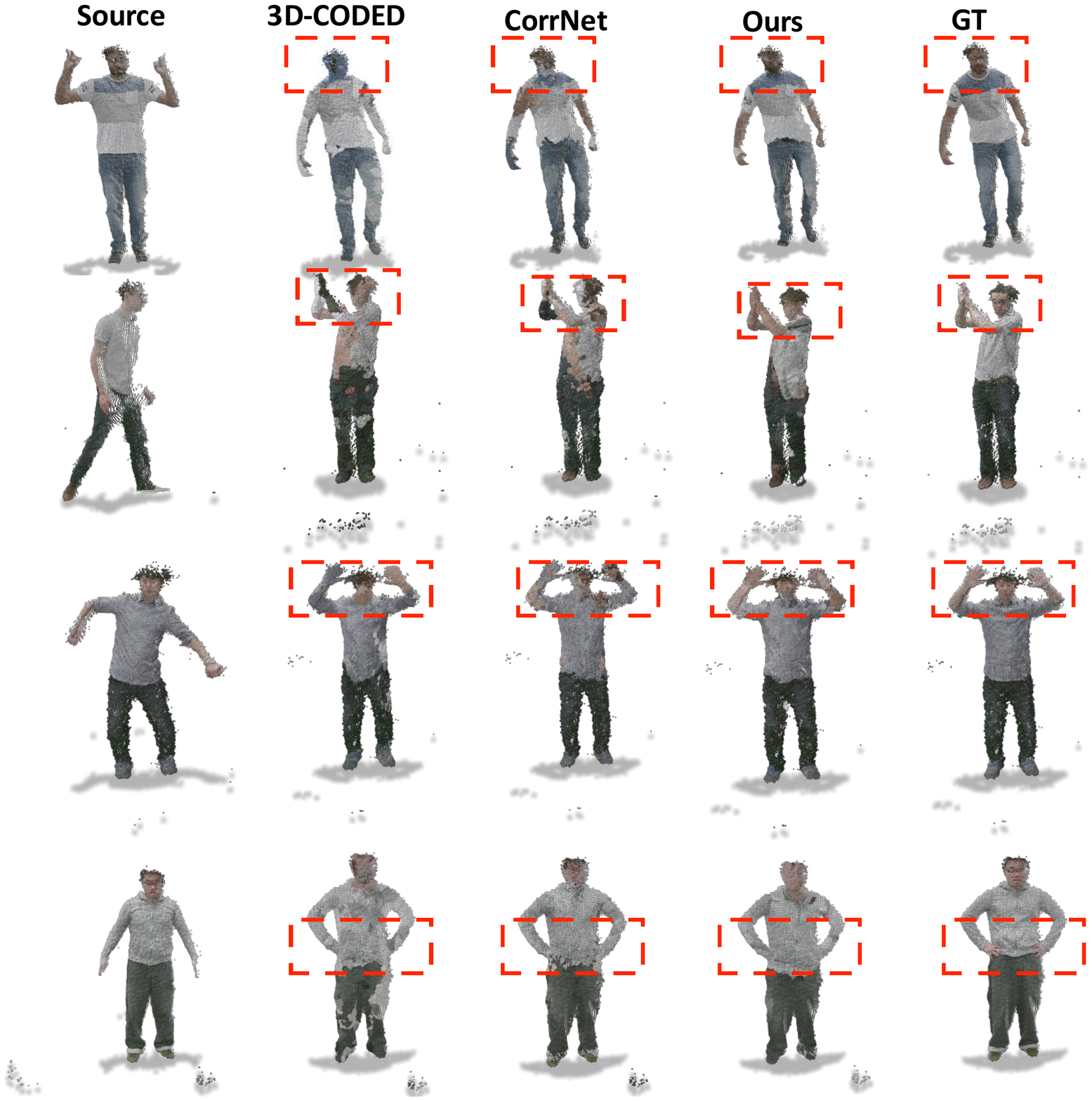}

\caption{Additional qualitative results on CMU Panoptic dataset through texture transfer. Stark differences are highlighted using a bounding box for better visualization. Last row depicts a failure case of our method.}
\label{fig:qual_additional_cmu}
\end{figure}

\subsection{Deforming Things 4D}
For the sake of completeness, we report additional qualitative results on \emph{inter-class} point clouds consisting of animals from the Deforming Things 4D dataset~\cite{li20214dcomplete}. This dataset consists of point clouds with self-occlusion and partiality, \emph{emulated} through Blender. Please note that unlike previous cases, there is no ground truth information available for inter-class shapes. For our qualitative example, we consider Cow, Bear, Fox and Deer classes. Our choice is based on large inter-class variability and non-isometry. All methods are trained on SMAL dataset~\cite{ZuffiCVPR2017} as mentioned in the main paper. 

\begin{figure}[H]
\includegraphics[width=1\linewidth]{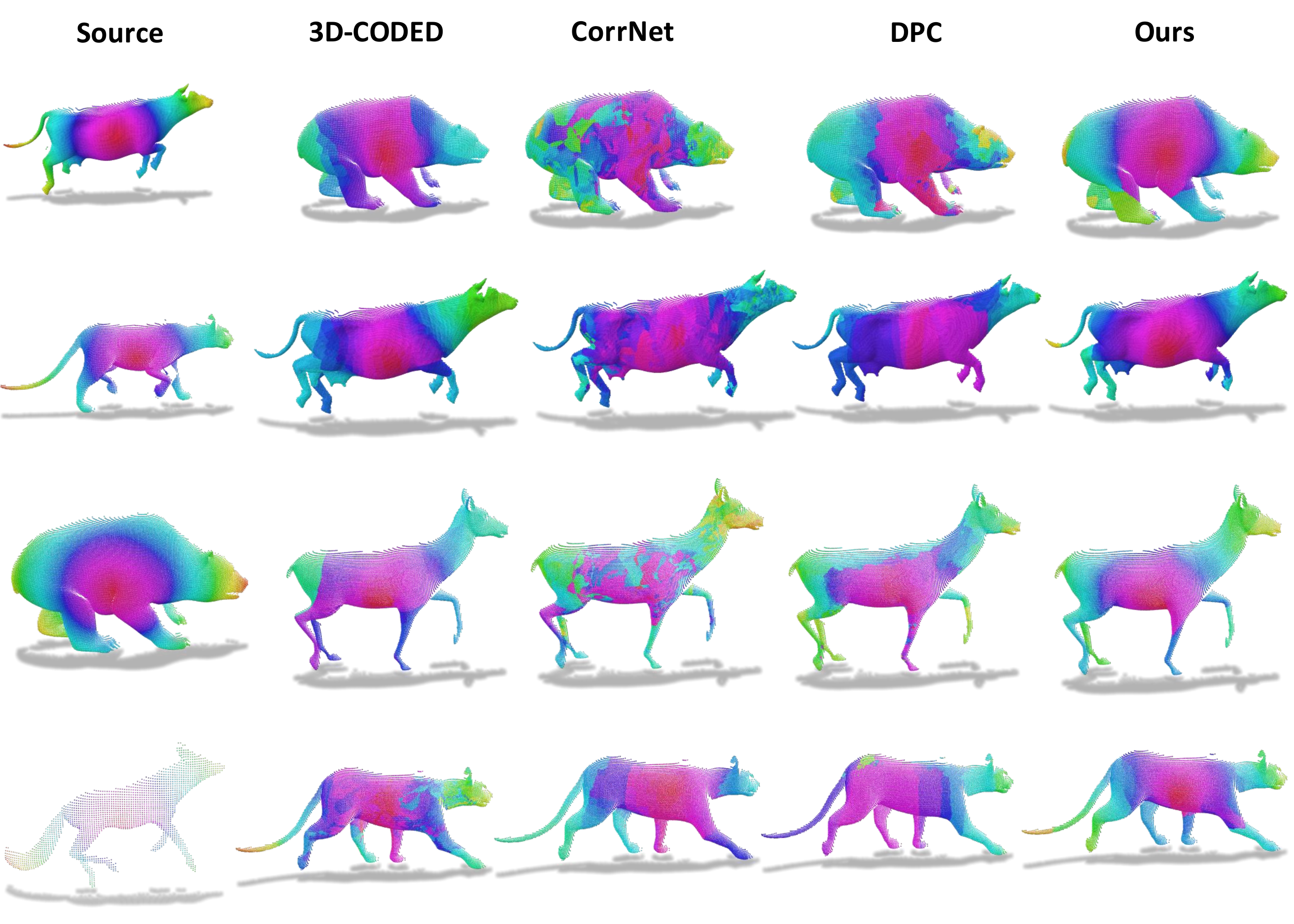}
\caption{Additional qualitative results on Deforming Things 4D animals dataset through color transfer. Our approach shows better qualitative correspondence for large non-isometry between point clouds. }
\label{fig:qual_DT4D}
\end{figure}

\subsection{Clothed humans : CAPE Scans}
\begin{figure}[H]
\includegraphics[width=1\linewidth]{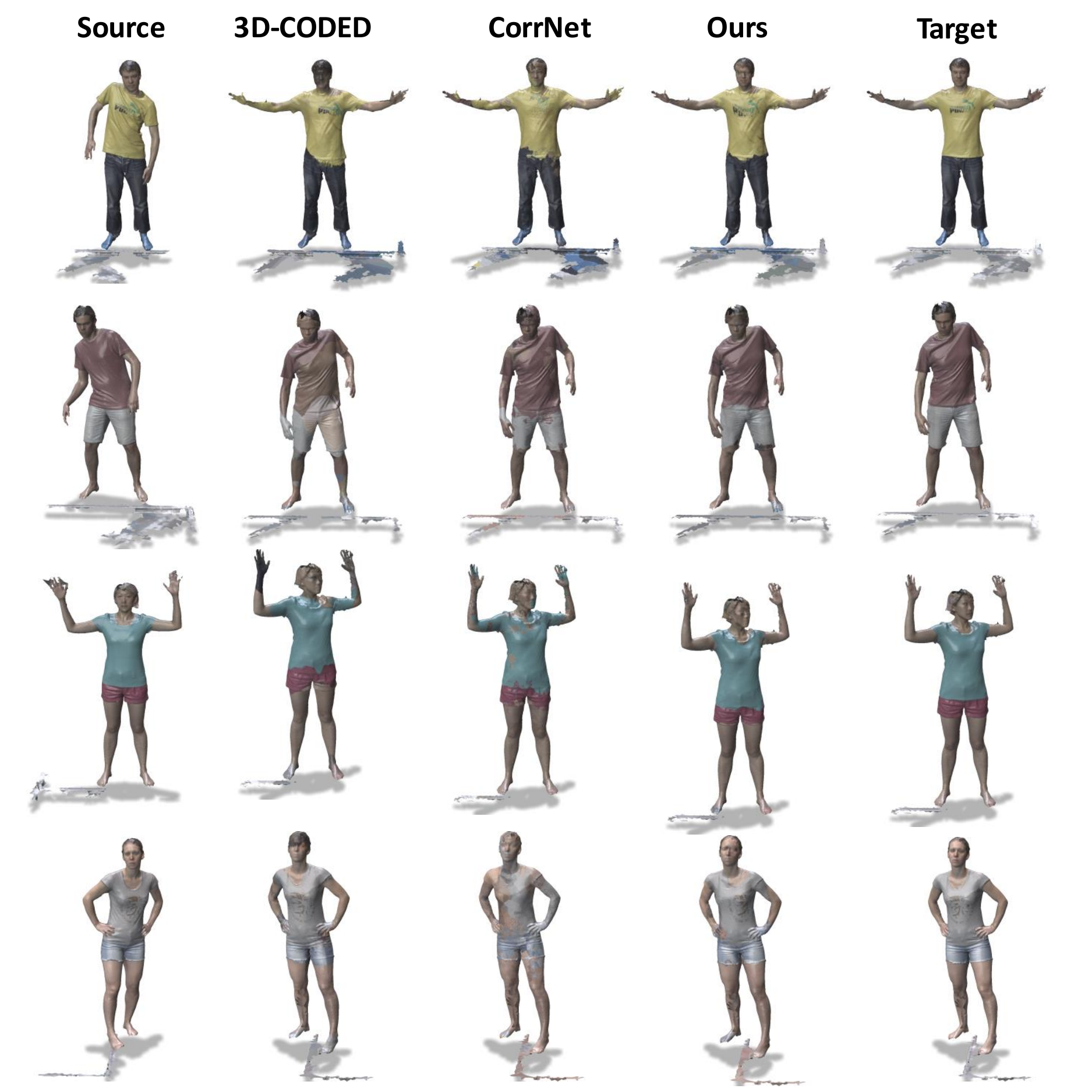}
\caption{Additional qualitative results on clothed humans from CAPE scans~\cite{CAPE:CVPR:20} consisting of noisy meshes with outliers. }  
\label{fig:qual_cape}
\end{figure}

\section{Limitations}
\label{sec:limitations}
While our method is largely robust through learning a volumetric map with strong regularisations, similar to all approaches that purely learn from extrinsic information, our approach suffers from generalization to unseen poses as depicted in the last row of Figure~\ref{fig:qual_additional_shrec19}. This issue can in part be attributed towards the ill-posed problem of learning an embedding space purely from Cartesian coordinates. However, our current framework of joint learning of latent spaces by continuous functions opens possibilities for descriptor learning alongside purely extrinsic information. Another notable failure case of our method occurs at the area of self-intersection as depicted in the last row of Figure~\ref{fig:qual_additional_cmu}. Making our approach robust to self-intersections is also an interesting future work.\\


\bibliographystyle{splncs04}
\bibliography{egbib}